\documentclass{article}

\usepackage{microtype}
\usepackage{graphicx}
\usepackage{subfigure}
\usepackage{booktabs} 

\usepackage{hyperref}

\usepackage[accepted]{icml2025}

\usepackage{amsmath}
\usepackage{amssymb}
\usepackage{mathtools}
\usepackage{amsthm}

\usepackage[capitalize,noabbrev]{cleveref}

\theoremstyle{plain}

\theoremstyle{definition}

\theoremstyle{remark}

\usepackage[textsize=tiny]{todonotes}

\definecolor{winter}{rgb}{0.85,0.08,0.2}
\definecolor{summer}{rgb}{0.95,0.53,0.18}         
\definecolor{spring}{rgb}{0.02,0.93,0.68}
\definecolor{autumn}{rgb}{0.02,0.68,0.9}

\icmltitlerunning{BAnG: Bidirectional Anchored Generation for Conditional RNA Design}

\begin{document}

\twocolumn[
\icmltitle{BAnG: Bidirectional Anchored Generation for Conditional RNA Design}



\icmlsetsymbol{equal}{*}

\begin{icmlauthorlist}
\icmlauthor{Roman Klypa}{gre}
\icmlauthor{Alberto Bietti}{fla}
\icmlauthor{Sergei Grudinin}{gre}
\end{icmlauthorlist}

\icmlaffiliation{gre}{ Univ. Grenoble Alpes, CNRS, Grenoble INP, LJK, 38000 Grenoble, France.}
\icmlaffiliation{fla}{Center
for Computational Mathematics, Flatiron Institute, 162 5th Ave, New York, NY 10010, USA}

\icmlcorrespondingauthor{Roman Klypa}{Roman.Klypa@univ-grenoble-alpes.fr}
\icmlcorrespondingauthor{Alberto Bietti}{Alberto.Bietti@gmail.com}
\icmlcorrespondingauthor{Sergei Grudinin}{Sergei.Grudinin@univ-grenoble-alpes.fr}

\icmlkeywords{generative modeling, sequential data, autoregressive generation, RNA design, biological sequences, deep learning models}

\vskip 0.3in
]



\printAffiliationsAndNotice{}  

\begin{abstract}

Designing RNA molecules that interact with specific proteins is a critical challenge in experimental and computational biology. Existing computational approaches require a substantial amount of previously known interacting RNA sequences for each specific protein or a detailed knowledge of RNA structure, restricting their utility in practice. To address this limitation, we develop RNA-BAnG, a deep learning-based model designed to generate RNA sequences for protein interactions without these requirements. Central to our approach is a novel generative method, Bidirectional Anchored Generation (BAnG), which leverages the observation that protein-binding RNA sequences often contain functional binding motifs embedded within broader sequence contexts. We first validate our method on generic synthetic tasks involving similar localized motifs to those appearing in RNAs, demonstrating its benefits over existing generative approaches. We then evaluate our model on biological sequences, showing its effectiveness for conditional RNA sequence design given a binding protein.

\end{abstract}

\section{Introduction}

Deep learning has significantly advanced bioinformatics and structural biology, particularly in predicting the structures, interactions, and functions of biomolecules \cite{callaway_chemistry_2024}.
It has also improved the efficiency of macromolecular design, facilitating applications in drug discovery and synthetic biology.
In particular, significant progress has been made in protein sequence design. Some illustrative examples include ESM3 \cite{hayes_simulating_2024} and Chroma \cite{ingraham_illuminating_2023}. This remarkable progress has not only revolutionized protein design but also opened new opportunities for addressing other complex biomolecular challenges. One such area is RNA generation, where similar principles of leveraging deep learning can be applied to advance our understanding and design of functional RNA~sequences.

Among the many challenges in RNA design, generating RNA sequences capable of binding to specific proteins stands out as a critical task with significant implications for understanding RNA-protein interactions  \cite{li_rna-protein_2024,fasogbon_recent_2024}. These interactions, central to essential biological processes such as gene regulation, splicing, and translation \cite{hentze_brave_2018}, highlight the need for precisely engineered RNA molecules. A notable example of such molecules are aptamers, short single-stranded RNA sequences that bind to specific proteins with high affinity and specificity. By acting as molecular inhibitors, probes, or delivery agents, aptamers offer versatile applications in therapeutics and diagnostics \cite{guo_engineering_2010, thavarajah_rna_2021}. Traditionally, aptamers are identified through SELEX (Systematic Evolution of Ligands by Exponential Enrichment), a labor-intensive experimental process. Developing computational methods to design aptamers could significantly accelerate and simplify their discovery, expanding their potential in biomedical applications.

Several studies have explored RNA generation in this domain. 
More classical approaches exploited evolutionary signals and statistical models \cite{kim_computational_2007,kim_ragpools_2007,aita_biomolecular_2010,tseng_entropic_2011,zhang_single-step_2023}, 
molecular modeling \cite{torkamanian-afshar_silico_2021}, 
and Monte Carlo tree search \cite{lee_predicting_2021, wang_discrete_2022, shin_aptatrans_2023, obonyo_rna_2024}.
More recent works used conditional variation autoencoders  \cite{chen_generating_2022,iwano_generative_2022,andress_daptev_2023}, long short-term memory models \cite{im_generative_2019, park_discovering_2020}, transformer-based architectures  \cite{zhao_generrna_2024, zhang_rnagenesis_2024}, and adversarial approach \cite{ozden_rnagen_2023}.
Most recent studies, e.g.,  AptaDiff \cite{wang_aptadiff_2024}, or RNAFLOW \cite{nori_rnaflow_2024} also explored diffusion processes and flow matching.
Almost all of the aforementioned approaches depend on a vast collection of nucleotide sequences known to interact with proteins to generate new ones, limiting their applicability to proteins for which extensive experimental data is available. To the best of our knowledge, RNAFLOW stands out as the only method that has not been trained on RNA affinity experimental data. However, it relies on RNA structure prediction tools to guide RNA design. Since these tools often lack the accuracy needed for precise RNA structures \cite{rhiju_nucleic_2024}, they ultimately reduce the model’s effectiveness in many practical scenarios.

In this work, we present a novel generative method combined with a deep-learning model that operates without relying on specific experimental data for the target protein and does not depend on RNA structural information. This approach enables a broader applicability and greater efficiency in RNA sequence generation than the ones mentioned~above.

The motivation for the design of the proposed generative method stems from two key observations. First, the total length of the RNA sequence to be generated is often unknown. Therefore, the method adopts an autoregressive approach. Second, RNA sequences that interact with proteins typically contain functional binding motifs -- specific regions that mediate interaction by forming molecular contacts with the protein. These binding motifs are embedded within larger sequence contexts, where the surrounding non-binding regions exert lesser influence on binding specificity \cite{ray_compendium_2013}. This makes it more effective to initiate sequence generation from the binding motif, rather than from the sequence's ends, as is commonly done in current state-of-the-art NLP autoregressive models. These are the core ideas behind our method, Bidirectional ANchored Generation (BAnG).

\begin{figure}[t]
\vskip 0.2in
\begin{center}
\centerline{\includegraphics[width=0.9\columnwidth]{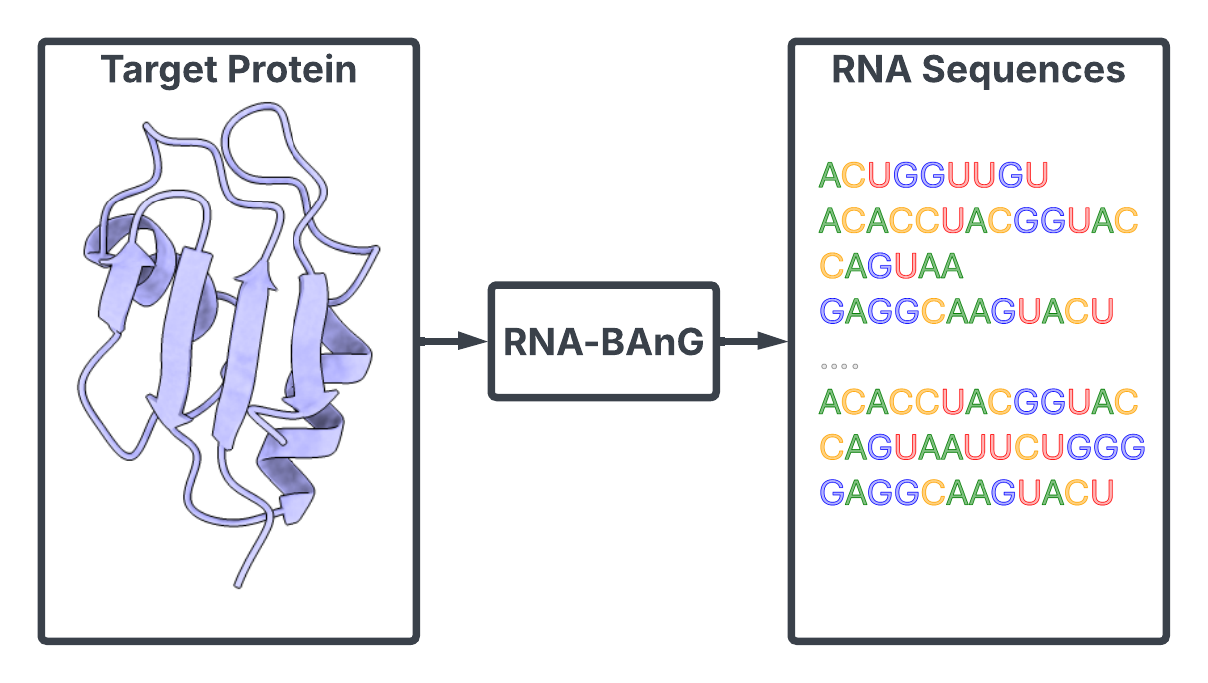}}
\caption{Schematic illustration of the RNA-BAnG generative process and its conditioning on the input protein 3D structure. The protein model was generated by AlphaFold2. RNA sequences are colored by nucleotides.
}
\label{fig:main}
\end{center}
\vskip -0.2in
\end{figure}

The model, named RNA-BAnG, is based on a transformer architecture with geometric attention. The latter allows for the incorporation of protein structural information, which is crucial for predicting RNA-protein interactions. By utilizing AlphaFold2 \cite{jumper_highly_2021}, a state-of-the-art protein structure prediction tool, we can obtain highly accurate structural data, even if the target protein is not solved experimentally.
The resulting combination of the RNA-BAnG model and the generative method, schematically illustrated in \cref{fig:main}, produces RNA sequences that interact with a given protein, utilizing both its sequence and structural information.
Our main contributions can be summarized as~follows:
\begin{enumerate}
    \item We propose a new bidirectional generation method, BAnG, along with a transformer-based architecture, RNA-BAnG, that are well-suited for RNA generation conditioned on a binding protein.
    \item We thoroughly validate the effectiveness of our method on relevant synthetic tasks and compare it with other widely used sequence generation methods.
    \item We evaluate our approach on experimental RNA-protein interaction data, showing promising results that outperform previous methods.
\end{enumerate}






 

 



 


\section{Bidirectional Anchored Generation for RNA}

In this section, we present our generative modeling approach, BAnG, and its application to conditional RNA sequence prediction through the RNA-BAnG model.
\subsection{Description of the BAnG Generative Approach}

\begin{figure}[tb]
\vskip 0.2in
\begin{center}
\centerline{\includegraphics[width=\columnwidth]{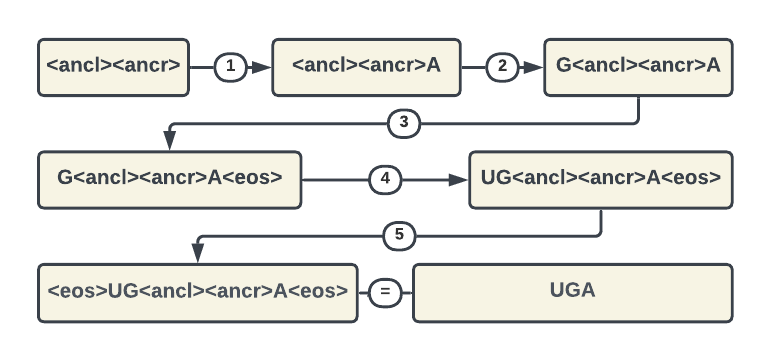}}
\caption{A step-by-step example process of RNA sequence 'UGA' generation.}
\label{fig:gen}
\end{center}
\vskip -0.2in
\end{figure}

Our approach models conditional distributions of each token given previously generated ones, similar to autoregressive modeling, but with a different factorization of the joint distribution over the sequence, which leads to a different order of generation.
In the BAnG framework, we leverage the following factorization of the joint distribution over a sequence $x = (x_{-m}, ..., x_0, ..., x_{n})$:
\begin{align}
P(x) = P(x_{0})  \underbrace{\prod_{i=1}^m P(x_{-i} | \mathcal{C}^l_{-i})}_{\text{left tokens}} 
 \underbrace{\prod_{i=1}^n P(x_{i} | \mathcal{C}^r_i)}_{\text{right tokens}},
\label{eq:main}
\end{align}
where $\mathcal{C}^l_{-i} = (x_{-i+1}, \dots, x_{i})$ and $\mathcal{C}^r_i = (x_{-i+1}, \dots, x_{i-1})$ denote left and right contexts. This formulation enables generation to proceed outward from the token $x_0$, conditioning each new token on a growing window of previously generated tokens. While the factorization is centered on $x_0$, we still need some initial tokens to begin generation. To this end, we insert two special anchor tokens, \texttt{<ancl>} and \texttt{<ancr>}, immediately before $x_0$: $x = (x_{-m}, ..., \texttt{<ancl>}, \texttt{<ancr>}, x_0, ..., x_{n})$. These tokens provide a fixed starting context, making it possible to assign a probability to $x_0$ and begin outward generation. They remain in the contexts $\mathcal{C}^l_{-i}$ and $\mathcal{C}^r_i$ for all subsequent tokens.

With this setup in place, generation proceeds token by token, alternating between the right and left directions: we first sample a token on the right, then a token on the left, and repeat this process to progressively extend the sequence outward. At each step, the distribution of the next token is conditioned on already generated ones, following Eq.~\ref{eq:main}. If an end-of-sequence (\texttt{<eos>}) token is generated in either direction, no further tokens are produced along that axis. The generation process stops when \texttt{<eos>} tokens are produced for both boundaries or when a predefined maximum sequence length is reached (see \cref{fig:gen}).

\begin{figure}[tb]
\vskip 0.2in
\begin{center}
\centerline{\includegraphics[width=0.8\columnwidth]{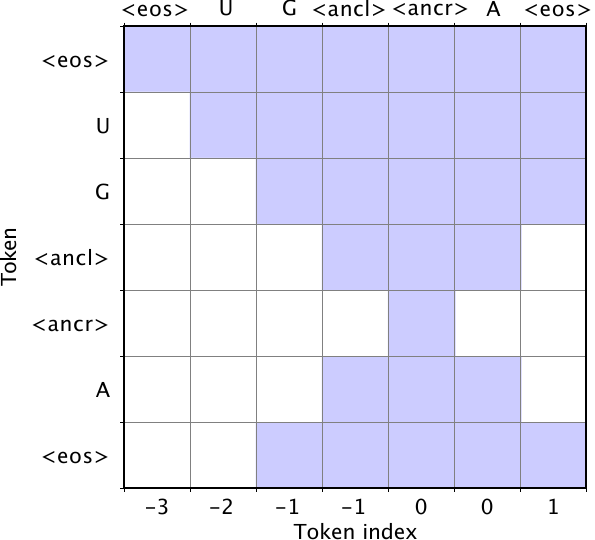}}
\caption{BAnG attention mask. Anchor tokens are indexed in a way to preserve relative distances in a sequence after anchors insertion.}
\label{fig:mask}
\end{center}
\vskip -0.2in
\end{figure}

\begin{figure}[t]
\vskip 0.2in
\begin{center}
\centerline{\includegraphics[width=\columnwidth]{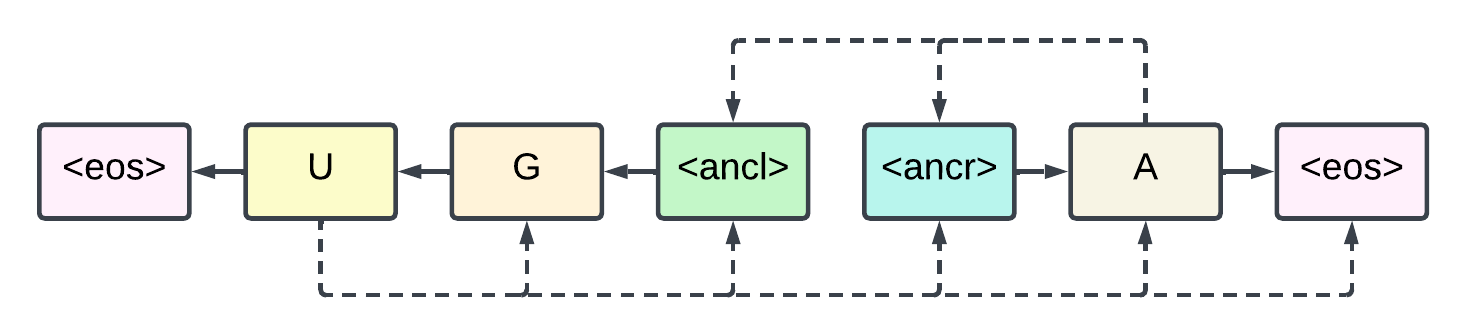}}
\caption{Schematic illustration of the masked attention mechanism and token probabilities derivation. Bold lines indicate the token embeddings from which the probabilities for each token are derived, while the dotted lines represent the tokens to which the given token's attention is directed.}
\label{fig:train}
\end{center}
\vskip -0.2in
\end{figure}

BAnG enables training a deep learning model to estimate the conditional distributions in Eq.~\ref{eq:main} in order to perform bidirectional generation in the described manner, using a process analogous to autoregressive training. The key difference in the architecture compared to the standard autoregressive case is the replacement of the conventional lower triangular attention mask with a specifically designed bidirectional attention mask, shown in \cref{fig:mask}. This custom mask ensures that any representation of a given token cannot depend on tokens beyond those appearing in the corresponding conditional in Eq.~\ref{eq:main}. This simple modification allows the model to learn the dependencies needed for our generation strategy, while ensuring efficient parallelization of the forward and backward passes across all tokens in a sequence during training. During model training and inference, the probabilities for the next token in each direction are derived from the embedding of the most recently generated token in the same direction, as shown schematically in \cref{fig:train}.

We shall emphasize that the single pass training with the conditional factorization in Eq.~\ref{eq:main} is only possible thanks to the introduction of two anchor tokens. Indeed, if only a single anchor token was used, it would be responsible for predicting the first tokens in both directions, right and left. To prevent information leakage, its attention would be restricted to itself, which would make the prediction of the left token independent of the right one. This lack of conditioning could lead to the generation of incompatible token pairs.

\subsection{Model}
\begin{figure}[t]
\vskip 0.2in
\begin{center}
\centerline{\includegraphics[width=1\columnwidth]{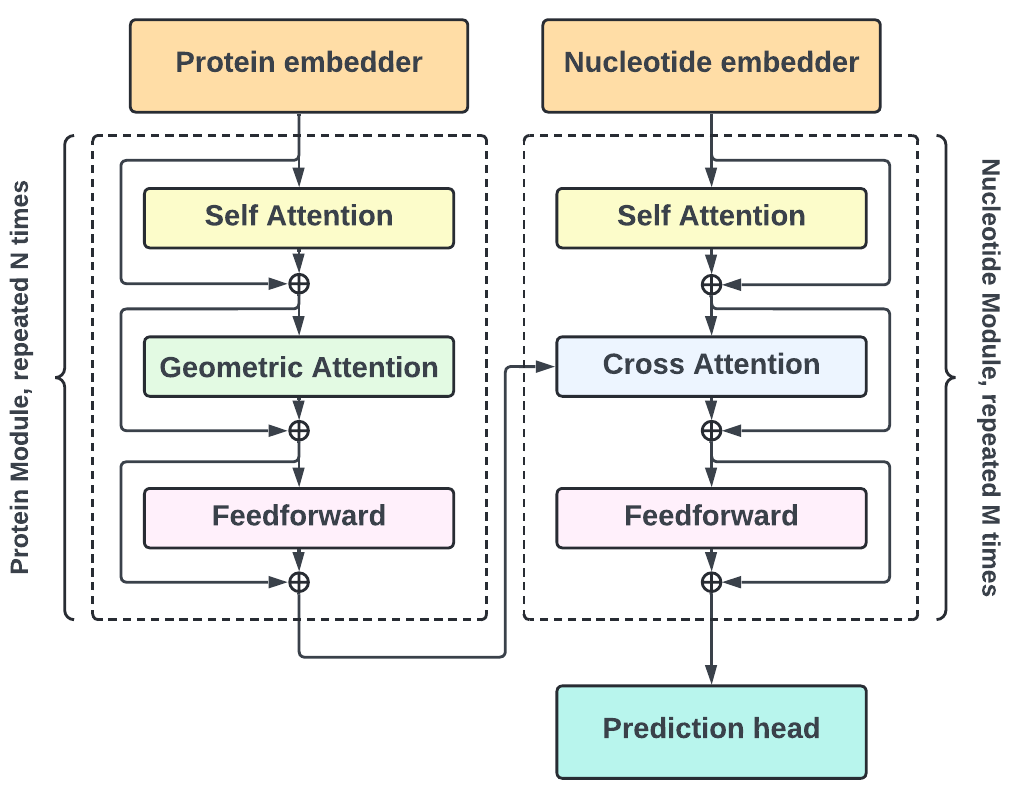}}
\caption{Schematic illustration of the RNA-BAnG architecture. The protein component is on the left, the nucleotide component is on the right. More details can be found in \cref{app:bang_model}.}
\label{fig:model}
\end{center}
\vskip -0.2in
\end{figure}

The RNA-BAnG architecture consists of two main components: a protein module and a nucleotide module. The protein module derives a representation from the protein's sequence and structure, while the nucleotide module generates a nucleotide sequence conditioned on this representation.
Our model's modules comprise several main blocks - {\bf Embedder}, {\bf Self Attention}, {\bf Geometric Attention}, and {\bf Cross Attention}, schematically illustrated in \cref{fig:model}.
The {\bf Embedder} block generates token embeddings from protein, RNA, and DNA sequences. The inclusion of RNA and DNA sequences in the training data serves to augment the dataset, as it can help the model learn the shared patterns.
Protein sequences are tokenized using the 20 standard amino acids and a padding token. RNA sequences are tokenized using the four canonical nucleotides, along with padding and end-of-sequence tokens. DNA sequences are tokenized with the same set, treating DNA residues as their RNA equivalents. To allow the model to distinguish between RNA and DNA, additional sequence type information is encoded separately (more details in \cref{app:bang_model}).

{\bf Self-attention} in the model was implemented in a classical way \cite{vaswani_attention_2023} using Rotary Position Embedding (RoPE) \cite{su_roformer_2023}. While RoPE has recently been predominantly used in autoregressive models, it was originally introduced for bidirectional transformers, making it particularly suitable for our case. Self-attention for protein sequences uses unmasked attention, while for nucleotide sequences, the BAnG mask is applied.

{\bf Cross-attention} is implemented similarly to the self-attention mechanism, with one key difference: instead of using RoPE, sinusoidal positional embeddings \cite{vaswani_attention_2023} are applied to the nucleotide sequence. Specifically, we encoded the positions of nucleotide tokens relative to the anchor ones, where the anchor token \texttt{<ancl>} is assigned an index of zero. This choice was made because residues in different chains do not have relative sequential distances, but we still want to include nucleotide position information in the calculation of its attention to the protein.

{\bf Geometric Attention} aims to incorporate protein structural information. To achieve this, we adopted the protein representation introduced in AlphaFold2 \cite{jumper_highly_2021}. Specifically, for each protein residue, a rigid frame $T$ is constructed based on the coordinates of its $C$, $C_{\alpha}$, and $N$ backbone atoms. Attention between these frames is then calculated using the geometric part from AlphaFold2's Invariant Point Attention (IPA) mechanism (more details in \cref{app:bang_model}).

We use GELU \cite{hendrycks_gaussian_2016} as the activation function in the Feed Forward layers. For normalization, we employed RMSNorm \cite{zhang_root_2019}. The number of attention heads is consistent across all attention blocks, and the dimension of each head is set independently of other model parameters. In both self-attention and cross-attention mechanisms, keys and queries are normalized prior to multiplication to prevent the attention-logit growth instability \cite{dehghani_scaling_2023}. The specific choice of hyperparameters is described in \cref{app:bang_model}.

\section{Validating BAnG on a Synthetic Task}

In this section, we evaluate the effectiveness of our BAnG strategy through a thorough analysis on a relevant synthetic task involving distinct subsequences.

\paragraph{Description of the task.}
To evaluate the BAnG method, we first consider a synthetic task that emulates a real-world conditional generation scenario. As mentioned above, RNA sequences often include a functional binding motif within a broader sequence context, where non-binding regions contribute minimally to binding specificity. The binding motifs are typically short and linear, with some being highly conserved and others exhibiting probabilistic patterns \cite{ray_compendium_2013}. Based on this observation, the task's objective is set to generate sequences that contain a predefined short subsequence, {\em synthetic motif}. Although these simplifications do not fully capture the nuances of real-world scenarios, they offer a baseline for understanding the method before applying it to more challenging tasks.

\paragraph{Synthetic data.}

The synthetic data consists of nucleotide sequences, each 50 residues long, with a synthetic motif placed at a random position. The remaining residues are uniformly distributed. We fixed two different random subsequences of length 6 as the synthetic motifs. The exact content of these motifs is not crucial to the task, as any possible subsequences of this length are equally likely to be uniformly sampled.  The length of six was chosen because it closely matches the size of real binding motifs \cite{ray_compendium_2013} and reduces the likelihood of random occurrences of such subsequences. As the anchor point for BAnG, we chose the center of the synthetic motif.

In the {\em SingleBind} training setup, sequences contained only the first synthetic motif, while in the {\em DoubleBind} setup, sequences could contain either the first or the second synthetic motif with equal probability. The objective of the SingleBind setup was to compare generative methods on a simpler task, while the goal of DoubleBind was to assess their performance under more realistic, thus more uncertain~conditions.

\begin{table}[t]
\caption{Comparison of generative methods on the SingleBind and DoubleBind tasks. Values represent the proportion of sequences that contain the correct synthetic motif.}
\label{table:sindual}
\vskip 0.15in
\begin{center}
\begin{small}
\begin{sc}
\begin{tabular}{lcccr}
\toprule
Generative method & SingleBind $\uparrow$ & DoubleBind  $\uparrow$\\
\midrule
BAnG    & {\bf 0.98} & {\bf 0.97} \\
autoregressive & 0.94 & 0.53 \\
IAnG entropy  & 0.91 & 0.54 \\
IAnG logit max  & 0.89 & 0.54 \\
iterative entropy  & 0.06 & 0.04 \\
iterative logit max  & 0.04 & 0.05 \\
random sequences    & 0.01 & 0.02  \\
\bottomrule
\end{tabular}
\end{sc}
\end{small}
\end{center}
\vskip -0.1in
\end{table}

\paragraph{Reference methods.}

We compare our method to existing generative approaches, including autoregressive generation, which is commonly used in natural language processing~\cite{bengio_neural_2000}, and iterative generation methods based on masked language modeling, as implemented in ESM3~\cite{hayes_simulating_2024}. Also, as the simplest baseline, we include in the comparison set of random sequences, where each token is sampled from a uniform distribution. For the autoregressive approach, we trained the model using a lower triangular attention mask. In the iterative approach, the model was trained with a demasking objective and a high masking rate of 50\%, which, according to the ESM3 authors, has been shown to yield effective generation results. Additional model training details for each tested method can be found in \cref{app:synth_model}. 

For the autoregressive approach, tokens are generated sequentially, one at a time, starting from the start-of-sequence (\texttt{<sos>}) token and continuing until the \texttt{<eos>} token is produced. In the iterative approach, however, all tokens but \texttt{<sos>} and \texttt{<eos>} are initially masked. Tokens are then unmasked one by one, with the next token to unmask chosen based on either the largest logit value (max logit decoding) or the smallest entropy (entropy decoding). Sampling details may be found in \cref{app:synth_sample}.

We also introduced a modification of iterative methods, better suited for the task - Iterative Anchored Generation (IAnG). This approach merges BAnG with iterative methods by incorporating an anchor token (\texttt{<anc>}) placed in the middle of the synthetic motif. In this method, the anchor token remains unmasked during training. At the start of the inference, the anchor token is positioned at a random location within the sequence and remains unmasked throughout the process.

\begin{table}[t]
\caption{Detailed statistics for the DoubleBind task: proportion of sequences containing either one of the synthetic motifs or both.}
\label{table:dual}
\vskip 0.15in
\begin{center}
\begin{small}
\begin{sc}
\begin{tabular}{lcccr}
\toprule
Generative method & First $\uparrow$ & Second $\uparrow$ & Both $\downarrow$ \\
\midrule
BAnG    & {\bf 0.43} &  {\bf 0.53} & 0.01 \\
autoregressive & 0.17 & 0.25 &  0.11 \\ 
IAnG logit max  & 0.09 & 0.44 & 0.01 \\
IAnG entropy  & 0.10 & 0.44 & 0.01 \\
iterative entropy  & 0.03 & 0.02 & 0.01 \\
iterative logit max & 0.01 & 0.04 & 0.01 \\
random sequences    & 0.01 & 0.01 & 0  \\
\bottomrule
\end{tabular}
\end{sc}
\end{small}
\end{center}
\vskip -0.1in
\end{table}

\paragraph{Evaluation results.}
With each tested approach, we generated 1,000 sequences. \cref{table:sindual} summarizes the performance of each method. Additional statistics on the frequency of each synthetic motif in the generated data for DoubleBind task are listed in \cref{table:dual}. Examples of generated sequences can be found in \cref{app:synth_seqs}.

The tables show that BAnG outperforms other methods, with a particularly notable margin on the DoubleBind task. BAnG also generates fewer sequences containing mixed synthetic motifs. We have also conducted experiments demonstrating BAnG's robustness to both larger fixed sequence lengths and varying sequence lengths, showing consistent performance across different configurations (\cref{app:synth_results}).

The very low performance of ESM3's iterative methods is expected, as the lack of absolute positional dependencies in the data causes the model to assign nearly uniform probabilities across tokens. This reasoning is further supported by the significant performance improvement observed when positional information is introduced through the anchor token in IAnG. Nonetheless, both methods suffer from a key limitation: a discrepancy between the demasking process during training and inference, which undermines their overall effectiveness.

Iterative methods may demonstrate improved performance in {\em modality translation}  scenarios, such as structure-to-sequence generation, which is a key focus of ESM3. However, they are unlikely to match the effectiveness of regressive approaches in purely generative tasks.

\section{Conditional RNA Generation}

In this section, we evaluate our RNA-BAnG modeling strategy for protein-conditioned generation of RNA sequences based on experimental biological data.

\subsection{Data}

We collected our protein-nucleotide interaction data from the Protein Data Bank (PDB) \cite{berman_protein_2000}, utilizing information provided in the PPI3D database \cite{dapkunas_ppi3d_2024}. However, we conducted independent postprocessing of the data, distinct from PPI3D, as they focus on structural RNA and DNA information, while we are concerned solely with their sequences. The postprocessing steps involved verifying chain interactions, discarding ambiguous protein structures, and excluding chains containing non-standard residues, as explained in more detail in \cref{app:data_double}.
During training and validation, each time a sample was encountered, its anchor point was randomly sampled from the interacting nucleotides. The anchor tokens were then inserted right after the selected nucleotide.
Additionally, to diversify RNA sequence information, we collected non-coding sequences from RNAcentral (release 24), a comprehensive database integrating RNA sequences from multiple expert sources \cite{thernacentralconsortium_rnacentral_2019} (more details in \cref{app:data_solo}). Since interaction data for these sequences is unavailable, their anchor points were selected randomly.

\subsection{RNA-BAnG Training}

We designed the training process to consist of two steps: the first for the model to learn general information about RNA sequences, and the second for the model to learn conditioning on proteins. In both steps, the training objective is the cross-entropy loss between the predicted and ground-truth nucleotide token probabilities \cite{bengio_neural_2000}. As the first step we train RNA-BAnG nucleotide module without cross-attention block on standalone RNA sequence data from RNAcentral. During training, the loss values for the four tokens closest to each anchor on either side are weighted at 0.01. This weighting scheme is applied because the model lacks the context needed to accurately predict these residues. Next, we train the full model, using weights from the previous step, on the combined protein-nucleotide sequence data, this time without applying any loss weighting. Additional training details can be found in \cref{app:train}.

\subsection{Evaluation Protocol}

Evaluating generative models is difficult because direct comparison with ground truth is not possible. To the best of our knowledge, there are no established computational protocols for measuring RNA affinity to proteins. Therefore, we use a predictive model as our main evaluation tool, following common practice in machine learning, such as FID scores in image generation or refoldability metrics in protein design.

\paragraph{Scoring model.} As the scoring model, we adopted DeepCLIP \cite{gronning_deepclip_2020}, following the approach of the authors of GenerRNA \cite{zhao_generrna_2024} and others \cite{im_generative_2019}. DeepCLIP is a state-of-the-art predictive model that, after being trained on examples of interacting and non-interacting RNA sequences for a given protein, can assign binding probabilities with the same protein to any RNA sequence. These probabilities serve as a proxy for evaluating the quality of sequences generated by RNA-BAnG and other methods.

\paragraph{Test set.} We built our test set using data from RNAcompete experiments \cite{ray_rapid_2009}, conducted by the authors of the RNA Compendium \cite{ray_compendium_2013}. Each sample consists of a protein sequence paired with a set of RNA sequences, each assigned a score reflecting its binding affinity to the protein. We trained a separate DeepCLIP model for each sample and excluded those where the scoring model did not meet our performance criteria (see \cref{app:eval_test}).

Due to the lack of experimentally solved protein structures for the test set and in pursuit of a more robust method, we used only AlphaFold2 protein models for RNA-BAnG inference during evaluation. For each sample in our test set, we generated the corresponding three-dimensional protein model and retained only those with a predicted local distance difference test (pLDDT) score greater than 70\%. This threshold, widely accepted in the community, ensures the structural reliability of the produced protein models, at least at the domain level. The final test set included 71 samples, representing 67 unique protein sequences.

To evaluate the generative models in a relative way, we built {\em positive} and {\em negative} sets of RNA sequences for each sample. The positive set includes sequences with the highest experimental affinity scores, while the negative set includes those with the lowest. We excluded these sets from DeepCLIP training and used them as an independent benchmark to assess both DeepCLIP predictions and the quality of the generated RNA sequences (see \cref{app:eval_test} for details). In addition to generating sequences with the evaluated model, we also created random sequences to serve as a baseline (see \cref{app:eval_sample}). Each mentioned set contains 1,000 sequences, unless stated otherwise.

\subsection{Experiment Results}

\begin{figure}[t]
\vskip 0.2in
\begin{center}
\centerline{\includegraphics[width=\columnwidth]{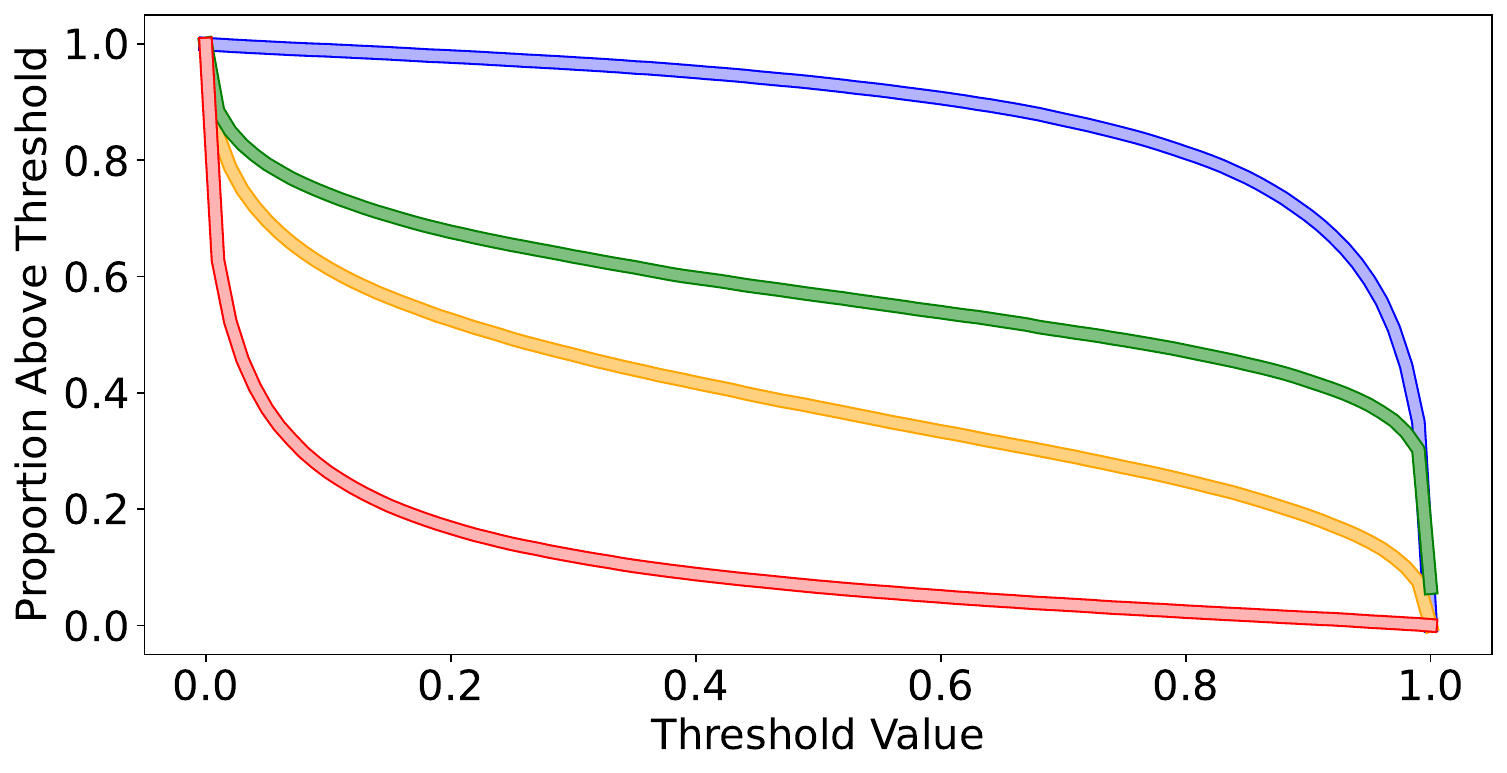}}
\caption{Proportion of sequences above the threshold: generated by RNA-BAnG (green) and randomly (yellow), the positive (blue) and negative (red) experimental sets. The values here represent the averages for the entire test set.
}
\label{fig:cdf}
\end{center}
\vskip -0.2in
\end{figure}

\begin{figure*}[t]
\vskip 0.2in
\begin{center}
\centerline{\includegraphics[width=\linewidth]{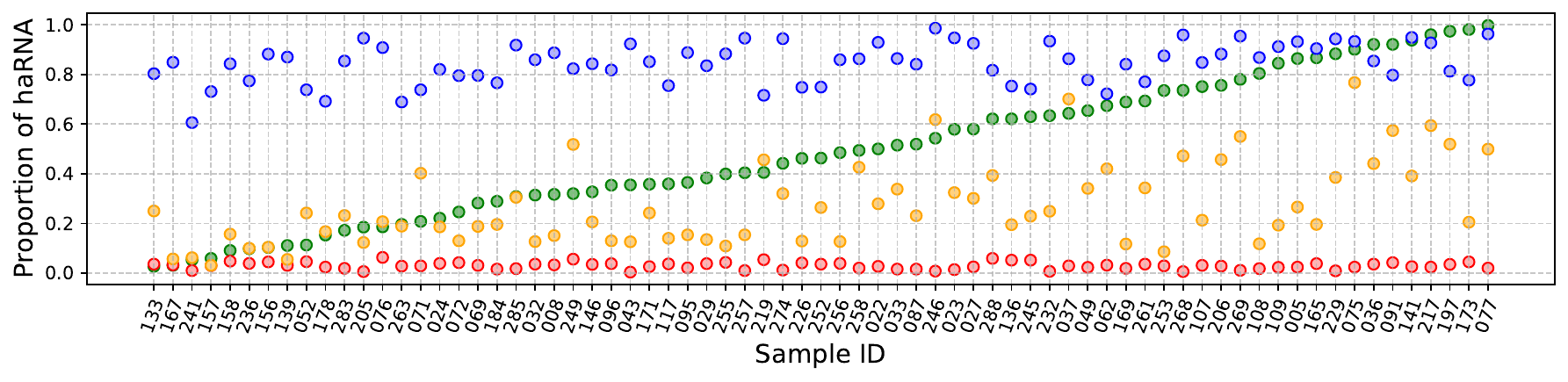}}
\caption{Proportions of high-affinity RNA sequences for each test sample, generated by RNA-BAnG (green) and randomly (yellow). These are compared with the proportions in the positive (blue) and negative (red) experimental sets. Test samples are ordered by RNA-BAnG performance. Sample IDs mapping to RNAcompete IDs mentioned in \cref{app:eval_test}.
}
\label{fig:res}
\end{center}
\vskip -0.2in
\end{figure*}

To estimate RNA-BAnG's performance, we use the proportion of sequences with DeepCLIP scores above a given threshold, rather than relying on the mean score for each test sample. This approach aligns better with practical applications, where the goal is often to maximize the number of sequences that meet a specific performance criterion, rather than optimizing for average performance across the entire~set.

\paragraph{Statistics of generated sequences.}

\cref{fig:cdf} presents the threshold-dependent performance curve, showing the relationship between sequences with DeepCLIP scores above a certain threshold and the threshold value. The area under the threshold-dependent performance curves in \cref{fig:cdf} captures the method's ability to generate affine sequences. A larger area indicates better performance, reflecting a higher proportion of sequences that exceed the threshold at different points. 
A high area value of 0.88 for the positive experimental set indicates that DeepCLIP is good at distinguishing high-affinity sequences from others. 
RNA-BAnG showed the value of 0.57, indicating moderate success in generating sequences that somewhat align with the positive set. 
The random set, with the area value of 0.39, suggests that our model is outperforming random generation. 
The negative set’s very low area value of 0.11 further underscores the DeepCLIP good performance, also suggesting that RNA-BAnG has some ability to avoid producing low-affinity outputs.

To complement the threshold-dependent performance analysis and capture variability concealed by averaging across samples, we conducted a sample-by-sample evaluation. Using a fixed threshold, we classified sequences as {\em high-affinity} RNA (haRNA) if their assigned score exceeded this value. For each sample in the test set, we then quantified the proportion of sequences meeting the haRNA criterion, presented in \cref{fig:res}. The fixed threshold value of 0.75 was chosen as detailed in the \cref{app:results}.
As one can see from \cref{fig:res}, RNA-BAnG generates more high-affinity sequences than a random generator for 56 out of 71 test samples. For 33 of the test samples, more than half of the sequences generated by RNA-BAnG are the high-affinity ones. Additionally, our model generated less than 20\% high-affinity sequences for only 14 out of 71 test samples.
The high success rate of randomly generated sequences for some samples may be attributed to the low complexity of captured protein binding motifs \cite{ray_compendium_2013}.

The performance of our model shows no significant correlation with the predicted quality of the protein 3D model (pLDDT) or with the sequence similarity of a protein target to the training data (\cref{fig:corr} in \cref{app:results}). This analysis suggests that the model does not only memorize the information, but exhibits a degree of generalization. We can also conclude that RNA-BAnG's performance is robust to the quality of protein structure prediction, provided the pLDDT score exceeds 70\%.

In addition to the deep learning-based scoring model, we explored several computational approaches to evaluate the generated sequences. Among these, one method yielded results that showed a meaningful correlation with the deep learning scores, supporting the validity of the model's output. Detailed descriptions of the computational analyses, along with examples of generated sequences for representative samples and individual DeepCLIP score distributions, are provided in \cref{app:results}.

\paragraph{Novelty and diversity of generated sequences.}
For our model's goal of accelerating experimental RNA design, diversity and novelty are essential metrics. By generating a wide range of diverse sequences, the model increases the chances of identifying optimal binders with varying affinities for the target protein, ensuring more effective candidates for experimental validation. Furthermore, the generation of novel sequences allows the discovery of unique RNA-protein interactions, which can lead to innovative therapeutic applications.

To quantify the diversity metric, we compute the ratio of the number of distinct clusters at a threshold of 0.9 sequence identity to the total number of generated sequences. A selected threshold of 0.9 is often considered as one of the lower limits used in RNA sequence clustering, especially when aiming to identify highly similar sequences or gene families \cite{edgar_updating_2018}. The resulting average diversity across the test set is 0.93 $\pm$ 0.13, indicating that the generated sequences are highly varied. The novelty is defined by the proportion of RNA sequences in the generated set that are not similar to the training data. We identify sequences for each test sample that have no similarity (see \cref{app:tools} for more details) to the training set, and the novelty is calculated as the ratio of these sequences to the total number of generated sequences. The resulting average novelty across the test set is 0.99 $\pm$ 0.01, indicating that the generated sequences are highly novel and distinct from the training data, which is a positive outcome for our model.

\begin{table}[t]
\caption{High-affinity RNA sequence proportion generated by GenerRNA and RNA-BAnG for each RNA Compendium sample. The protein sequences are identical for the same genes. Although underlined samples do not meet our test set selection criteria, they are included to enhance the diversity of the comparison. Sample IDs mapping to RNAcompete IDs mentioned in \cref{app:eval_test}.}
\label{table:generrna}
\vskip 0.15in
\begin{center}
\begin{small}
\begin{sc}
\begin{tabular}{lccccr}
\toprule
Sample ID & Gene  & GenerRNA & RNA-BAnG \\
\midrule
\underline{106}   & SRSF1 & 0.42 & {\bf 0.49} \\
107    & SRSF1 & 0.59 & {\bf 0.74}  \\
108    & SRSF1 & 0.56  & {\bf 0.77}  \\
109    & SRSF1 & 0.62  & {\bf 0.85}  \\
\underline{110}    & SRSF1 & 0.43  & {\bf 0.66}  \\
\underline{121}    & ELAVL1 & 0.64  & {\bf 0.91} \\
\bottomrule
\end{tabular}
\end{sc}
\end{small}
\end{center}
\vskip -0.1in
\end{table}

\paragraph{Comparison with similar models.} We compare our model, RNA-BAnG, with two other methods, GenerRNA \cite{zhao_generrna_2024} and RNAFLOW \cite{nori_rnaflow_2024}, as they are the most relevant existing methods for generating RNA sequences for a diverse set of proteins.
The first model, GenerRNA, leverages a substantial collection of RNA sequences known to interact with a specific protein during its fine-tuning process, allowing it to generate additional sequences with similar binding properties. 
Consequently, our comparison is limited to the proteins for which GenerRNA was fine-tuned by its authors. We used their published inference results, removing generated RNA larger than 50 nucleotides, leaving 921 and 909 sequences for the ELAVL1 and SRSF1 proteins, respectively. 
These proteins are already present in the RNA Compendium samples. As shown in \cref{table:generrna}, our model generates more high-affinity RNA sequences than GenerRNA. It is important to note that RNA-BAnG generates these sequences without relying on any additional information, whereas GenerRNA required extensive data mining from multiple experimental studies.

\begin{figure}[t]
\vskip 0.2in
\begin{center}
\centerline{\includegraphics[width=\columnwidth]{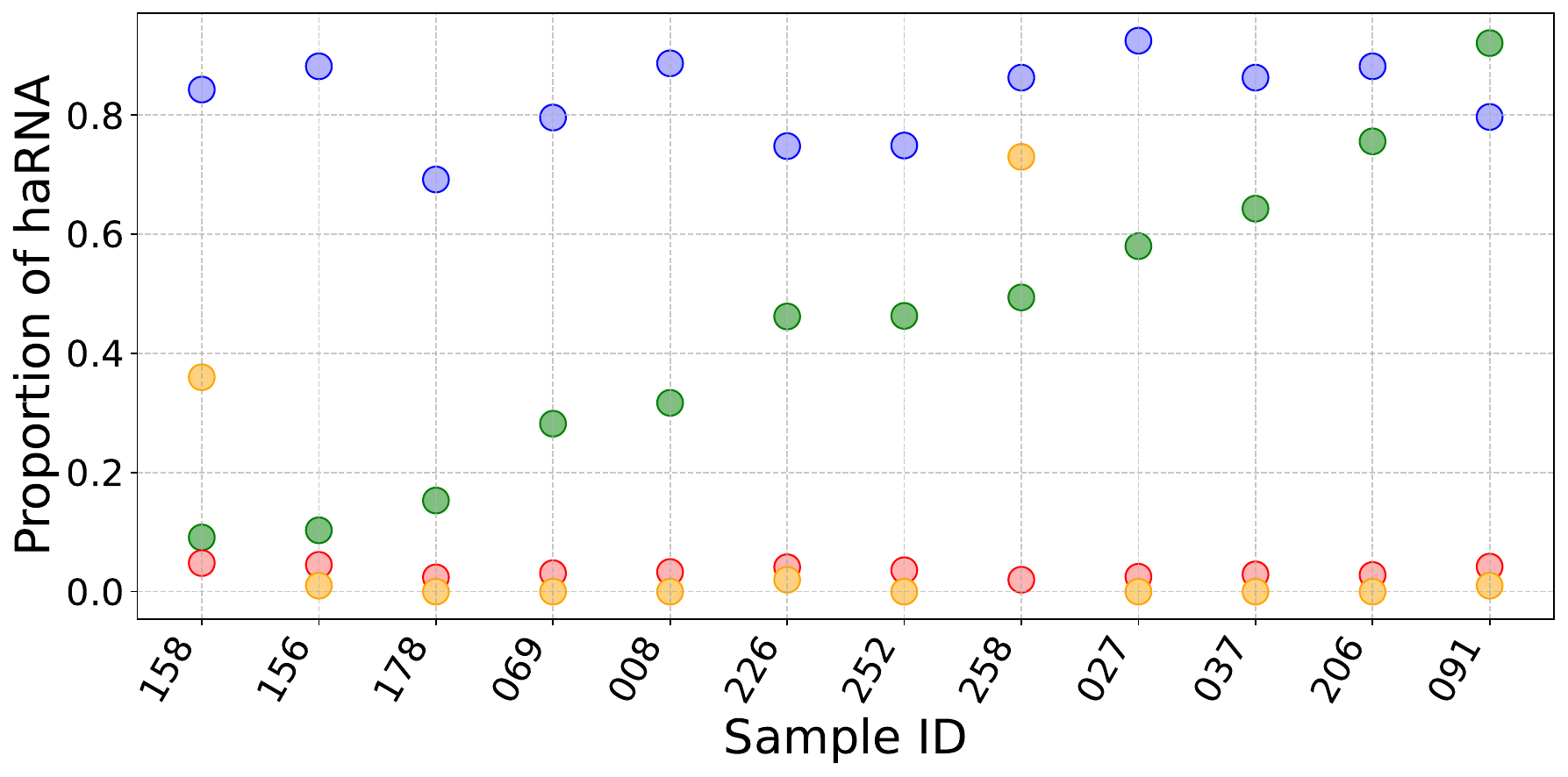}}
\caption{Proportion of high affinity sequences generated by RNA-BAnG (green) and RNAFLOW (yellow).
Samples are sorted by the performance of RNA-BAnG. Sample IDs mapping to RNAcompete IDs mentioned in \cref{app:eval_test}.}
\label{fig:rnaflow}
\end{center}
\vskip -0.2in
\end{figure}

The second baseline model is RNAFLOW, which uses RNA structure prediction tools. Due to its much lower speed, approximately 50 times slower than RNA-BAnG, we only generated RNA sequences for proteins that had zero sequence similarity with our training set, resulting in 100 sequences of length 50 for each protein. 
Unfortunately, RNAFLOW generated sequences with an unusually high frequences of G and C nucleotides — 62\% and 32\%, respectively — regardless of protein structure or sequence. As a result, it produced high-affinity sequences for only a couple of proteins, and none for the others. Consequently, RNA-BAnG outperformed RNAFLOW on most of the test samples (\cref{fig:rnaflow}).

The low performance of RNAFLOW can be explained by a discrepancy between its training process and the inference we conducted. Concretely, RNAFLOW was trained on protein sequences and structures truncated to 50 residues, ensuring the inclusion of the protein binding site. However, we tested it on proteins of varying lengths, ranging from 177 to 589 amino acids, without highlighting their binding sites through truncation (since this information is unavailable). Moreover, the proteins we tested had zero sequence similarity to those used to train RNA-BAnG. As both RNA-BAnG and RNAFLOW are trained on PDB-derived data, it is reasonable to assume low similarity to RNAFLOW’s training set as well. This could make prediction of their structure in complex with RNA more challenging. These limitations significantly restrict the applicability of RNAFLOW to RNA sequence design, especially when compared to RNA-BAnG.

\section{Conclusion}

This study introduces a novel deep-learning model, RNA-BAnG, and a sequence generation method, BAnG, for designing RNA sequences that bind to a given protein. Unlike previous approaches, our model demonstrates remarkable flexibility by eliminating the need for extensive structural or interaction data during inference. Although our model relies on protein structure, AlphaFold predictions make sequence information sufficient. This innovation significantly broadens the applicability of our method, making it a versatile tool for RNA-protein interaction studies.

Our method is based on the observation that RNA sequences contain functional binding motifs, while the surrounding sequence context is less critical for interaction. When evaluated on a synthetic task against other sequence generation approaches, BAnG demonstrated superior performance. Importantly, the method's design makes it applicable beyond RNA-protein interactions, extending to any scenario where the focus is on optimizing functional subsequences within a larger sequence.
According to the state-of-the-art DeepCLIP scoring, RNA-BAnG outperforms existing methods, generating a higher proportion of sequences with strong predicted binding affinity. The generated sequences exhibit both diversity and novelty, expanding the range of potential RNA candidates for further experimental validation.

Future work could focus on integrating experimental feedback to further refine the model, optimizing its architecture for enhanced performance, and improving its usability for broader applications. Additionally, experimental validation of the generated sequences would provide further insights into the practical applicability of our method.
In summary, our work represents a significant step forward in RNA sequence generation, offering a powerful and flexible tool for researchers in the field of RNA-protein interactions.

\paragraph{Software and Data.}
The code 
and the model, along with the model weights, are available at \url{https://github.com/rsklypa/RNA-BAnG}. 
The code allows for running the RNA sequences generation process for any correct protein 3D structure as input.

\section*{Impact Statement}

The main purpose of this work is to advance the field of generative models.
However, the applications of this method may have social and industrial benefits. Potential applications include in-silico SELEX approaches, RNA vaccine design, the development of novel drugs, and some other pharmaceutical tasks.

\section*{Acknowledgments}

This work was granted access to the HPC resources of IDRIS under the allocation 2024-AD011015647 made by GENCI. A substantial part of the computations presented in this paper was performed using the GRICAD infrastructure (https://gricad.univ-grenoble-alpes.fr), which is supported by Grenoble research communities. The authors are very thankful to
Yoann Fauconnet and Jessica Andreani from I2BC, CEA Paris-Saclay
for their support and advise on RNA scoring models.
The authors also acknowledge the support from the PHC MAIMONIDE 2023 program, project number 49285YJ.

\bibliography{main}

\begin{thebibliography}{53}
\providecommand{\natexlab}[1]{#1}
\providecommand{\url}[1]{\texttt{#1}}
\expandafter\ifx\csname urlstyle\endcsname\relax
  \providecommand{\doi}[1]{doi: #1}\else
  \providecommand{\doi}{doi: \begingroup \urlstyle{rm}\Url}\fi

\bibitem[Aita \& Husimi(2010)Aita and Husimi]{aita_biomolecular_2010}
Aita, T. and Husimi, Y.
\newblock Biomolecular information gained through in vitro evolution.
\newblock \emph{Biophysical reviews}, 2:\penalty0 1--11, 2010.
\newblock Publisher: Springer.

\bibitem[Altschul et~al.(1990)Altschul, Gish, Miller, Myers, and Lipman]{altschul_basic_1990}
Altschul, S.~F., Gish, W., Miller, W., Myers, E.~W., and Lipman, D.~J.
\newblock Basic local alignment search tool.
\newblock \emph{Journal of Molecular Biology}, 215\penalty0 (3):\penalty0 403--410, October 1990.
\newblock ISSN 00222836.
\newblock \doi{10.1016/S0022-2836(05)80360-2}.
\newblock URL \url{https://linkinghub.elsevier.com/retrieve/pii/S0022283605803602}.

\bibitem[Andress et~al.(2023)Andress, Kappel, Villena, Cuperlovic-Culf, Yan, and Li]{andress_daptev_2023}
Andress, C., Kappel, K., Villena, M.~E., Cuperlovic-Culf, M., Yan, H., and Li, Y.
\newblock {DAPTEV}: {Deep} aptamer evolutionary modelling for {COVID}-19 drug design.
\newblock \emph{PLOS Computational Biology}, 19\penalty0 (7):\penalty0 e1010774, July 2023.
\newblock ISSN 1553-7358.
\newblock \doi{10.1371/journal.pcbi.1010774}.
\newblock URL \url{https://dx.plos.org/10.1371/journal.pcbi.1010774}.

\bibitem[Bengio et~al.(2000)Bengio, Ducharme, and Vincent]{bengio_neural_2000}
Bengio, Y., Ducharme, R., and Vincent, P.
\newblock A {Neural} {Probabilistic} {Language} {Model}.
\newblock In Leen, T., Dietterich, T., and Tresp, V. (eds.), \emph{Advances in {Neural} {Information} {Processing} {Systems}}, volume~13. MIT Press, 2000.
\newblock URL \url{https://proceedings.neurips.cc/paper_files/paper/2000/file/728f206c2a01bf572b5940d7d9a8fa4c-Paper.pdf}.

\bibitem[Berman(2000)]{berman_protein_2000}
Berman, H.~M.
\newblock The {Protein} {Data} {Bank}.
\newblock \emph{Nucleic Acids Research}, 28\penalty0 (1):\penalty0 235--242, January 2000.
\newblock ISSN 13624962.
\newblock \doi{10.1093/nar/28.1.235}.
\newblock URL \url{https://academic.oup.com/nar/article-lookup/doi/10.1093/nar/28.1.235}.

\bibitem[Callaway(2024)]{callaway_chemistry_2024}
Callaway, E.
\newblock Chemistry {Nobel} goes to developers of {AlphaFold} {AI} that predicts protein structures.
\newblock \emph{Nature}, 634\penalty0 (8034):\penalty0 525--526, October 2024.
\newblock ISSN 0028-0836, 1476-4687.
\newblock \doi{10.1038/d41586-024-03214-7}.
\newblock URL \url{https://www.nature.com/articles/d41586-024-03214-7}.

\bibitem[Chen et~al.(2022)Chen, Chen, Shen, Wornow, Bae, Yeh, Hsu, and Liu]{chen_generating_2022}
Chen, J.~C., Chen, J.~P., Shen, M.~W., Wornow, M., Bae, M., Yeh, W.-H., Hsu, A., and Liu, D.~R.
\newblock Generating experimentally unrelated target molecule-binding highly functionalized nucleic-acid polymers using machine learning.
\newblock \emph{Nature Communications}, 13\penalty0 (1):\penalty0 4541, August 2022.
\newblock ISSN 2041-1723.
\newblock \doi{10.1038/s41467-022-31955-4}.
\newblock URL \url{https://www.nature.com/articles/s41467-022-31955-4}.

\bibitem[Dapkūnas et~al.(2024)Dapkūnas, Timinskas, Olechnovič, Tomkuvienė, and Venclovas]{dapkunas_ppi3d_2024}
Dapkūnas, J., Timinskas, A., Olechnovič, K., Tomkuvienė, M., and Venclovas, C.
\newblock {PPI3D}: a web server for searching, analyzing and modeling protein–protein, protein–peptide and protein–nucleic acid interactions.
\newblock \emph{Nucleic Acids Research}, 52\penalty0 (W1):\penalty0 W264--W271, July 2024.
\newblock ISSN 0305-1048, 1362-4962.
\newblock \doi{10.1093/nar/gkae278}.
\newblock URL \url{https://academic.oup.com/nar/article/52/W1/W264/7645776}.

\bibitem[Dehghani et~al.(2023)Dehghani, Djolonga, Mustafa, Padlewski, Heek, Gilmer, Steiner, Caron, Geirhos, Alabdulmohsin, Jenatton, Beyer, Tschannen, Arnab, Wang, Riquelme, Minderer, Puigcerver, Evci, Kumar, Steenkiste, Elsayed, Mahendran, Yu, Oliver, Huot, Bastings, Collier, Gritsenko, Birodkar, Vasconcelos, Tay, Mensink, Kolesnikov, Pavetić, Tran, Kipf, Lučić, Zhai, Keysers, Harmsen, and Houlsby]{dehghani_scaling_2023}
Dehghani, M., Djolonga, J., Mustafa, B., Padlewski, P., Heek, J., Gilmer, J., Steiner, A., Caron, M., Geirhos, R., Alabdulmohsin, I., Jenatton, R., Beyer, L., Tschannen, M., Arnab, A., Wang, X., Riquelme, C., Minderer, M., Puigcerver, J., Evci, U., Kumar, M., Steenkiste, S.~v., Elsayed, G.~F., Mahendran, A., Yu, F., Oliver, A., Huot, F., Bastings, J., Collier, M.~P., Gritsenko, A., Birodkar, V., Vasconcelos, C., Tay, Y., Mensink, T., Kolesnikov, A., Pavetić, F., Tran, D., Kipf, T., Lučić, M., Zhai, X., Keysers, D., Harmsen, J., and Houlsby, N.
\newblock Scaling {Vision} {Transformers} to 22 {Billion} {Parameters}, February 2023.
\newblock URL \url{http://arxiv.org/abs/2302.05442}.
\newblock arXiv:2302.05442 [cs].

\bibitem[Edgar(2018)]{edgar_updating_2018}
Edgar, R.~C.
\newblock Updating the 97\% identity threshold for {16S} ribosomal {RNA} {OTUs}.
\newblock \emph{Bioinformatics}, 34\penalty0 (14):\penalty0 2371--2375, July 2018.
\newblock ISSN 1367-4803, 1367-4811.
\newblock \doi{10.1093/bioinformatics/bty113}.
\newblock URL \url{https://academic.oup.com/bioinformatics/article/34/14/2371/4913809}.

\bibitem[Fasogbon et~al.(2024)Fasogbon, Ondari, Tusubira, Rangasamy, Venkatesan, Musyoka, and Aja]{fasogbon_recent_2024}
Fasogbon, I.~V., Ondari, E.~N., Tusubira, D., Rangasamy, L., Venkatesan, J., Musyoka, A.~M., and Aja, P.~M.
\newblock {RECENT} {FOCUS} {IN} {NON}-{SELEX}-{COMPUTATIONAL} {APPROACH} {FOR} {DE} {NOVO} {APTAMER} {DESIGN}: {A} {MINI} {REVIEW}.
\newblock \emph{Analytical Biochemistry}, pp.\  115756, 2024.
\newblock Publisher: Elsevier.

\bibitem[Grønning et~al.(2020)Grønning, Doktor, Larsen, Petersen, Holm, Bruun, Hansen, Hartung, Baumbach, and Andresen]{gronning_deepclip_2020}
Grønning, A., Doktor, T.~K., Larsen, S., Petersen, U., Holm, L.~L., Bruun, G., Hansen, M.~B., Hartung, A.-M., Baumbach, J., and Andresen, B.~S.
\newblock {DeepCLIP}: predicting the effect of mutations on protein–{RNA} binding with deep learning.
\newblock \emph{Nucleic Acids Research}, pp.\  gkaa530, June 2020.
\newblock ISSN 0305-1048, 1362-4962.
\newblock \doi{10.1093/nar/gkaa530}.
\newblock URL \url{https://academic.oup.com/nar/advance-article/doi/10.1093/nar/gkaa530/5859960}.

\bibitem[Guo et~al.(2010)Guo, Coban, Snead, Trebley, Hoeprich, Guo, and Shu]{guo_engineering_2010}
Guo, P., Coban, O., Snead, N.~M., Trebley, J., Hoeprich, S., Guo, S., and Shu, Y.
\newblock Engineering {RNA} for {Targeted} {siRNA} {Delivery} and {Medical} {Application}.
\newblock \emph{Advanced Drug Delivery Reviews}, 62\penalty0 (6):\penalty0 650--666, April 2010.
\newblock ISSN 0169409X.
\newblock \doi{10.1016/j.addr.2010.03.008}.
\newblock URL \url{https://linkinghub.elsevier.com/retrieve/pii/S0169409X10000773}.

\bibitem[Hayes et~al.(2024)Hayes, Rao, Akin, Sofroniew, Oktay, Lin, Verkuil, Tran, Deaton, Wiggert, Badkundri, Shafkat, Gong, Derry, Molina, Thomas, Khan, Mishra, Kim, Bartie, Nemeth, Hsu, Sercu, Candido, and Rives]{hayes_simulating_2024}
Hayes, T., Rao, R., Akin, H., Sofroniew, N.~J., Oktay, D., Lin, Z., Verkuil, R., Tran, V.~Q., Deaton, J., Wiggert, M., Badkundri, R., Shafkat, I., Gong, J., Derry, A., Molina, R.~S., Thomas, N., Khan, Y., Mishra, C., Kim, C., Bartie, L.~J., Nemeth, M., Hsu, P.~D., Sercu, T., Candido, S., and Rives, A.
\newblock Simulating 500 million years of evolution with a language model, July 2024.
\newblock URL \url{http://biorxiv.org/lookup/doi/10.1101/2024.07.01.600583}.

\bibitem[Hendrycks \& Gimpel(2016)Hendrycks and Gimpel]{hendrycks_gaussian_2016}
Hendrycks, D. and Gimpel, K.
\newblock Gaussian {Error} {Linear} {Units} ({GELUs}), 2016.
\newblock URL \url{https://arxiv.org/abs/1606.08415}.
\newblock Version Number: 5.

\bibitem[Hentze et~al.(2018)Hentze, Castello, Schwarzl, and Preiss]{hentze_brave_2018}
Hentze, M.~W., Castello, A., Schwarzl, T., and Preiss, T.
\newblock A brave new world of {RNA}-binding proteins.
\newblock \emph{Nature Reviews Molecular Cell Biology}, 19\penalty0 (5):\penalty0 327--341, May 2018.
\newblock ISSN 1471-0072, 1471-0080.
\newblock \doi{10.1038/nrm.2017.130}.
\newblock URL \url{https://www.nature.com/articles/nrm.2017.130}.

\bibitem[Im et~al.(2019)Im, Park, and Han]{im_generative_2019}
Im, J., Park, B., and Han, K.
\newblock A generative model for constructing nucleic acid sequences binding to a protein.
\newblock \emph{BMC genomics}, 20\penalty0 (Suppl 13):\penalty0 967, 2019.
\newblock Publisher: Springer.

\bibitem[Ingraham et~al.(2023)Ingraham, Baranov, Costello, Barber, Wang, Ismail, Frappier, Lord, Ng-Thow-Hing, Van~Vlack, Tie, Xue, Cowles, Leung, Rodrigues, Morales-Perez, Ayoub, Green, Puentes, Oplinger, Panwar, Obermeyer, Root, Beam, Poelwijk, and Grigoryan]{ingraham_illuminating_2023}
Ingraham, J.~B., Baranov, M., Costello, Z., Barber, K.~W., Wang, W., Ismail, A., Frappier, V., Lord, D.~M., Ng-Thow-Hing, C., Van~Vlack, E.~R., Tie, S., Xue, V., Cowles, S.~C., Leung, A., Rodrigues, J.~V., Morales-Perez, C.~L., Ayoub, A.~M., Green, R., Puentes, K., Oplinger, F., Panwar, N.~V., Obermeyer, F., Root, A.~R., Beam, A.~L., Poelwijk, F.~J., and Grigoryan, G.
\newblock Illuminating protein space with a programmable generative model.
\newblock \emph{Nature}, 623\penalty0 (7989):\penalty0 1070--1078, November 2023.
\newblock ISSN 0028-0836, 1476-4687.
\newblock \doi{10.1038/s41586-023-06728-8}.
\newblock URL \url{https://www.nature.com/articles/s41586-023-06728-8}.

\bibitem[Iwano et~al.(2022)Iwano, Adachi, Aoki, Nakamura, and Hamada]{iwano_generative_2022}
Iwano, N., Adachi, T., Aoki, K., Nakamura, Y., and Hamada, M.
\newblock Generative aptamer discovery using {RaptGen}.
\newblock \emph{Nature Computational Science}, 2\penalty0 (6):\penalty0 378--386, June 2022.
\newblock ISSN 2662-8457.
\newblock \doi{10.1038/s43588-022-00249-6}.
\newblock URL \url{https://www.nature.com/articles/s43588-022-00249-6}.

\bibitem[Jumper et~al.(2021)Jumper, Evans, Pritzel, Green, Figurnov, Ronneberger, Tunyasuvunakool, Bates, Žídek, Potapenko, Bridgland, Meyer, Kohl, Ballard, Cowie, Romera-Paredes, Nikolov, Jain, Adler, Back, Petersen, Reiman, Clancy, Zielinski, Steinegger, Pacholska, Berghammer, Bodenstein, Silver, Vinyals, Senior, Kavukcuoglu, Kohli, and Hassabis]{jumper_highly_2021}
Jumper, J., Evans, R., Pritzel, A., Green, T., Figurnov, M., Ronneberger, O., Tunyasuvunakool, K., Bates, R., Žídek, A., Potapenko, A., Bridgland, A., Meyer, C., Kohl, S. A.~A., Ballard, A.~J., Cowie, A., Romera-Paredes, B., Nikolov, S., Jain, R., Adler, J., Back, T., Petersen, S., Reiman, D., Clancy, E., Zielinski, M., Steinegger, M., Pacholska, M., Berghammer, T., Bodenstein, S., Silver, D., Vinyals, O., Senior, A.~W., Kavukcuoglu, K., Kohli, P., and Hassabis, D.
\newblock Highly accurate protein structure prediction with {AlphaFold}.
\newblock \emph{Nature}, 596\penalty0 (7873):\penalty0 583--589, August 2021.
\newblock ISSN 0028-0836, 1476-4687.
\newblock \doi{10.1038/s41586-021-03819-2}.
\newblock URL \url{https://www.nature.com/articles/s41586-021-03819-2}.

\bibitem[Katoh \& Standley(2013)Katoh and Standley]{katoh_mafft_2013}
Katoh, K. and Standley, D.~M.
\newblock {MAFFT} {Multiple} {Sequence} {Alignment} {Software} {Version} 7: {Improvements} in {Performance} and {Usability}.
\newblock \emph{Molecular Biology and Evolution}, 30\penalty0 (4):\penalty0 772--780, April 2013.
\newblock ISSN 0737-4038, 1537-1719.
\newblock \doi{10.1093/molbev/mst010}.
\newblock URL \url{https://academic.oup.com/mbe/article-lookup/doi/10.1093/molbev/mst010}.

\bibitem[Kim et~al.(2007{\natexlab{a}})Kim, Gan, and Schlick]{kim_computational_2007}
Kim, N., Gan, H.~H., and Schlick, T.
\newblock A computational proposal for designing structured {RNA} pools for in vitro selection of {RNAs}.
\newblock \emph{RNA}, 13\penalty0 (4):\penalty0 478--492, 2007{\natexlab{a}}.
\newblock Publisher: Cold Spring Harbor Lab.

\bibitem[Kim et~al.(2007{\natexlab{b}})Kim, Shin, Elmetwaly, Gan, and Schlick]{kim_ragpools_2007}
Kim, N., Shin, J.~S., Elmetwaly, S., Gan, H.~H., and Schlick, T.
\newblock {RagPools}: {RNA}-{As}-{Graph}-{Pools}: a web server for assisting the design of structured {RNA} pools for in vitro selection.
\newblock \emph{Bioinformatics}, 23\penalty0 (21):\penalty0 2959--2960, 2007{\natexlab{b}}.
\newblock Publisher: Oxford University Press.

\bibitem[Kingma \& Ba(2017)Kingma and Ba]{kingma_adam_2017}
Kingma, D.~P. and Ba, J.
\newblock Adam: {A} {Method} for {Stochastic} {Optimization}, January 2017.
\newblock URL \url{http://arxiv.org/abs/1412.6980}.
\newblock arXiv:1412.6980 [cs].

\bibitem[Korhonen et~al.(2017)Korhonen, Palin, Taipale, and Ukkonen]{korhonen_fast_2017}
Korhonen, J.~H., Palin, K., Taipale, J., and Ukkonen, E.
\newblock Fast motif matching revisited: high-order {PWMs}, {SNPs} and indels.
\newblock \emph{Bioinformatics}, 33\penalty0 (4):\penalty0 514--521, February 2017.
\newblock ISSN 1367-4803, 1367-4811.
\newblock \doi{10.1093/bioinformatics/btw683}.
\newblock URL \url{https://academic.oup.com/bioinformatics/article/33/4/514/2726114}.

\bibitem[Lee et~al.(2021)Lee, Jang, Kang, and Song]{lee_predicting_2021}
Lee, G., Jang, G.~H., Kang, H.~Y., and Song, G.
\newblock Predicting aptamer sequences that interact with target proteins using an aptamer-protein interaction classifier and a {Monte} {Carlo} tree search approach.
\newblock \emph{PloS one}, 16\penalty0 (6):\penalty0 e0253760, 2021.
\newblock Publisher: Public Library of Science San Francisco, CA USA.

\bibitem[Li et~al.(2024)Li, Huang, Cui, Towey, Zhou, Tian, and Zou]{li_rna-protein_2024}
Li, D., Huang, R., Cui, C., Towey, D., Zhou, L., Tian, J., and Zou, B.
\newblock {RNA}-{Protein} {Interaction} {Prediction} {Based} on {Deep} {Learning}: {A} {Comprehensive} {Survey}.
\newblock \emph{arXiv preprint arXiv:2410.00077}, 2024.

\bibitem[Li \& Godzik(2006)Li and Godzik]{li_cd-hit_2006}
Li, W. and Godzik, A.
\newblock Cd-hit: a fast program for clustering and comparing large sets of protein or nucleotide sequences.
\newblock \emph{Bioinformatics}, 22\penalty0 (13):\penalty0 1658--1659, July 2006.
\newblock ISSN 1367-4811, 1367-4803.
\newblock \doi{10.1093/bioinformatics/btl158}.
\newblock URL \url{https://academic.oup.com/bioinformatics/article/22/13/1658/194225}.

\bibitem[Nori \& Jin(2024)Nori and Jin]{nori_rnaflow_2024}
Nori, D. and Jin, W.
\newblock {RNAFlow}: {RNA} {Structure} \& {Sequence} {Design} via {Inverse} {Folding}-{Based} {Flow} {Matching}, June 2024.
\newblock URL \url{http://arxiv.org/abs/2405.18768}.
\newblock arXiv:2405.18768 [q-bio].

\bibitem[Obonyo et~al.(2024)Obonyo, Jouandeau, and Owuor]{obonyo_rna_2024}
Obonyo, S., Jouandeau, N., and Owuor, D.
\newblock {RNA} {Generative} {Modeling} {With} {Tree} {Search}.
\newblock In \emph{2024 {IEEE} {Conference} on {Computational} {Intelligence} in {Bioinformatics} and {Computational} {Biology} ({CIBCB})}, pp.\  1--9. IEEE, 2024.

\bibitem[Olechnovič \& Venclovas(2014)Olechnovič and Venclovas]{olechnovic_voronota_2014}
Olechnovič, K. and Venclovas, C.
\newblock Voronota: {A} fast and reliable tool for computing the vertices of the {Voronoi} diagram of atomic balls.
\newblock \emph{Journal of Computational Chemistry}, 35\penalty0 (8):\penalty0 672--681, March 2014.
\newblock ISSN 0192-8651, 1096-987X.
\newblock \doi{10.1002/jcc.23538}.
\newblock URL \url{https://onlinelibrary.wiley.com/doi/10.1002/jcc.23538}.

\bibitem[Olechnovič \& Venclovas(2021)Olechnovič and Venclovas]{olechnovic_vorocontacts_2021}
Olechnovič, K. and Venclovas, C.
\newblock {VoroContacts}: a tool for the analysis of interatomic contacts in macromolecular structures.
\newblock \emph{Bioinformatics}, 37\penalty0 (24):\penalty0 4873--4875, December 2021.
\newblock ISSN 1367-4803, 1367-4811.
\newblock \doi{10.1093/bioinformatics/btab448}.
\newblock URL \url{https://academic.oup.com/bioinformatics/article/37/24/4873/6300513}.

\bibitem[Ozden et~al.(2023)Ozden, Barazandeh, Akboga, Tabrizi, Seker, and Cicek]{ozden_rnagen_2023}
Ozden, F., Barazandeh, S., Akboga, D., Tabrizi, S.~S., Seker, U. O.~S., and Cicek, A.~E.
\newblock {RNAGEN}: {A} generative adversarial network-based model to generate synthetic {RNA} sequences to target proteins, July 2023.
\newblock URL \url{http://biorxiv.org/lookup/doi/10.1101/2023.07.11.548246}.

\bibitem[Park \& Han(2020)Park and Han]{park_discovering_2020}
Park, B. and Han, K.
\newblock Discovering protein-binding {RNA} motifs with a generative model of {RNA} sequences.
\newblock \emph{Computational Biology and Chemistry}, 84:\penalty0 107171, February 2020.
\newblock ISSN 14769271.
\newblock \doi{10.1016/j.compbiolchem.2019.107171}.
\newblock URL \url{https://linkinghub.elsevier.com/retrieve/pii/S1476927119305365}.

\bibitem[Ray et~al.(2009)Ray, Kazan, Chan, Castillo, Chaudhry, Talukder, Blencowe, Morris, and Hughes]{ray_rapid_2009}
Ray, D., Kazan, H., Chan, E.~T., Castillo, L.~P., Chaudhry, S., Talukder, S., Blencowe, B.~J., Morris, Q., and Hughes, T.~R.
\newblock Rapid and systematic analysis of the {RNA} recognition specificities of {RNA}-binding proteins.
\newblock \emph{Nature Biotechnology}, 27\penalty0 (7):\penalty0 667--670, July 2009.
\newblock ISSN 1087-0156, 1546-1696.
\newblock \doi{10.1038/nbt.1550}.
\newblock URL \url{https://www.nature.com/articles/nbt.1550}.

\bibitem[Ray et~al.(2013)Ray, Kazan, Cook, Weirauch, Najafabadi, Li, Gueroussov, Albu, Zheng, Yang, Na, Irimia, Matzat, Dale, Smith, Yarosh, Kelly, Nabet, Mecenas, Li, Laishram, Qiao, Lipshitz, Piano, Corbett, Carstens, Frey, Anderson, Lynch, Penalva, Lei, Fraser, Blencowe, Morris, and Hughes]{ray_compendium_2013}
Ray, D., Kazan, H., Cook, K.~B., Weirauch, M.~T., Najafabadi, H.~S., Li, X., Gueroussov, S., Albu, M., Zheng, H., Yang, A., Na, H., Irimia, M., Matzat, L.~H., Dale, R.~K., Smith, S.~A., Yarosh, C.~A., Kelly, S.~M., Nabet, B., Mecenas, D., Li, W., Laishram, R.~S., Qiao, M., Lipshitz, H.~D., Piano, F., Corbett, A.~H., Carstens, R.~P., Frey, B.~J., Anderson, R.~A., Lynch, K.~W., Penalva, L. O.~F., Lei, E.~P., Fraser, A.~G., Blencowe, B.~J., Morris, Q.~D., and Hughes, T.~R.
\newblock A compendium of {RNA}-binding motifs for decoding gene regulation.
\newblock \emph{Nature}, 499\penalty0 (7457):\penalty0 172--177, July 2013.
\newblock ISSN 0028-0836, 1476-4687.
\newblock \doi{10.1038/nature12311}.
\newblock URL \url{https://www.nature.com/articles/nature12311}.

\bibitem[Reddi et~al.(2019)Reddi, Kale, and Kumar]{reddi_convergence_2019}
Reddi, S.~J., Kale, S., and Kumar, S.
\newblock On the {Convergence} of {Adam} and {Beyond}, April 2019.
\newblock URL \url{http://arxiv.org/abs/1904.09237}.
\newblock arXiv:1904.09237 [cs].

\bibitem[Rhiju et~al.(2024)Rhiju, Shujun, Alissa, and Rachael]{rhiju_nucleic_2024}
Rhiju, D., Shujun, H., Alissa, H., and Rachael, K.
\newblock Nucleic {Acid} {Assessment} {CASP16}, December 2024.
\newblock URL \url{https://predictioncenter.org/casp16/doc/presentations/Day-3/Day3-01-Kretsch_CASP16_NA_Assessement_PuntaCana_RCK_v1.pptx}.

\bibitem[Schneider \& Stephens(1990)Schneider and Stephens]{schneider_sequence_1990}
Schneider, T.~D. and Stephens, R.
\newblock Sequence logos: a new way to display consensus sequences.
\newblock \emph{Nucleic Acids Research}, 18\penalty0 (20):\penalty0 6097--6100, 1990.
\newblock ISSN 0305-1048, 1362-4962.
\newblock \doi{10.1093/nar/18.20.6097}.
\newblock URL \url{https://academic.oup.com/nar/article-lookup/doi/10.1093/nar/18.20.6097}.

\bibitem[Shin et~al.(2023)Shin, Kang, Kim, Sel, Choi, Lee, Kang, and Song]{shin_aptatrans_2023}
Shin, I., Kang, K., Kim, J., Sel, S., Choi, J., Lee, J.-W., Kang, H.~Y., and Song, G.
\newblock {AptaTrans}: a deep neural network for predicting aptamer-protein interaction using pretrained encoders.
\newblock \emph{BMC bioinformatics}, 24\penalty0 (1):\penalty0 447, 2023.
\newblock Publisher: Springer.

\bibitem[Stormo et~al.(1982)Stormo, Schneider, Gold, and Ehrenfeucht]{stormo_use_1982}
Stormo, G.~D., Schneider, T.~D., Gold, L., and Ehrenfeucht, A.
\newblock Use of the ‘{Perceptron}’ algorithm to distinguish translational initiation sites in \textit{{E}. coli}.
\newblock \emph{Nucleic Acids Research}, 10\penalty0 (9):\penalty0 2997--3011, 1982.
\newblock ISSN 0305-1048, 1362-4962.
\newblock \doi{10.1093/nar/10.9.2997}.
\newblock URL \url{https://academic.oup.com/nar/article-lookup/doi/10.1093/nar/10.9.2997}.

\bibitem[Su et~al.(2023)Su, Lu, Pan, Murtadha, Wen, and Liu]{su_roformer_2023}
Su, J., Lu, Y., Pan, S., Murtadha, A., Wen, B., and Liu, Y.
\newblock {RoFormer}: {Enhanced} {Transformer} with {Rotary} {Position} {Embedding}, November 2023.
\newblock URL \url{http://arxiv.org/abs/2104.09864}.
\newblock arXiv:2104.09864 [cs].

\bibitem[Thavarajah et~al.(2021)Thavarajah, Hertz, Bushhouse, Archuleta, and Lucks]{thavarajah_rna_2021}
Thavarajah, W., Hertz, L.~M., Bushhouse, D.~Z., Archuleta, C.~M., and Lucks, J.~B.
\newblock {RNA} {Engineering} for {Public} {Health}: {Innovations} in {RNA}-{Based} {Diagnostics} and {Therapeutics}.
\newblock \emph{Annual Review of Chemical and Biomolecular Engineering}, 12\penalty0 (1):\penalty0 263--286, June 2021.
\newblock ISSN 1947-5438, 1947-5446.
\newblock \doi{10.1146/annurev-chembioeng-101420-014055}.
\newblock URL \url{https://www.annualreviews.org/doi/10.1146/annurev-chembioeng-101420-014055}.

\bibitem[{The RNAcentral Consortium} et~al.(2019){The RNAcentral Consortium}, Sweeney, Petrov, Burkov, Finn, Bateman, Szymanski, Karlowski, Gorodkin, Seemann, Cannone, Gutell, Fey, Basu, Kay, Cochrane, Billis, Emmert, Marygold, Huntley, Lovering, Frankish, Chan, Lowe, Bruford, Seal, Vandesompele, Volders, Paraskevopoulou, Ma, Zhang, Griffiths-Jones, Bujnicki, Boccaletto, Blake, Bult, Chen, Zhao, Wood, Rutherford, Rivas, Cole, Laulederkind, Shimoyama, Gillespie, Orlic-Milacic, Kalvari, Nawrocki, Engel, Cherry, Team, Berardini, Hatzigeorgiou, Karagkouni, Howe, Davis, Dinger, He, Yoshihama, Kenmochi, Stadler, and Williams]{thernacentralconsortium_rnacentral_2019}
{The RNAcentral Consortium}, Sweeney, B.~A., Petrov, A.~I., Burkov, B., Finn, R.~D., Bateman, A., Szymanski, M., Karlowski, W.~M., Gorodkin, J., Seemann, S.~E., Cannone, J.~J., Gutell, R.~R., Fey, P., Basu, S., Kay, S., Cochrane, G., Billis, K., Emmert, D., Marygold, S.~J., Huntley, R.~P., Lovering, R.~C., Frankish, A., Chan, P.~P., Lowe, T.~M., Bruford, E., Seal, R., Vandesompele, J., Volders, P.-J., Paraskevopoulou, M., Ma, L., Zhang, Z., Griffiths-Jones, S., Bujnicki, J.~M., Boccaletto, P., Blake, J.~A., Bult, C.~J., Chen, R., Zhao, Y., Wood, V., Rutherford, K., Rivas, E., Cole, J., Laulederkind, S. J.~F., Shimoyama, M., Gillespie, M.~E., Orlic-Milacic, M., Kalvari, I., Nawrocki, E., Engel, S.~R., Cherry, J.~M., Team, S., Berardini, T.~Z., Hatzigeorgiou, A., Karagkouni, D., Howe, K., Davis, P., Dinger, M., He, S., Yoshihama, M., Kenmochi, N., Stadler, P.~F., and Williams, K.~P.
\newblock {RNAcentral}: a hub of information for non-coding {RNA} sequences.
\newblock \emph{Nucleic Acids Research}, 47\penalty0 (D1):\penalty0 D221--D229, January 2019.
\newblock ISSN 0305-1048, 1362-4962.
\newblock \doi{10.1093/nar/gky1034}.
\newblock URL \url{https://academic.oup.com/nar/article/47/D1/D221/5160993}.

\bibitem[Torkamanian-Afshar et~al.(2021)Torkamanian-Afshar, Nematzadeh, Tabarzad, Najafi, Lanjanian, and Masoudi-Nejad]{torkamanian-afshar_silico_2021}
Torkamanian-Afshar, M., Nematzadeh, S., Tabarzad, M., Najafi, A., Lanjanian, H., and Masoudi-Nejad, A.
\newblock In silico design of novel aptamers utilizing a hybrid method of machine learning and genetic algorithm.
\newblock \emph{Molecular diversity}, 25:\penalty0 1395--1407, 2021.
\newblock Publisher: Springer.

\bibitem[Tseng et~al.(2011)Tseng, Ashrafuzzaman, Mane, Kapty, Mercer, and Tuszynski]{tseng_entropic_2011}
Tseng, C.-Y., Ashrafuzzaman, M., Mane, J.~Y., Kapty, J., Mercer, J.~R., and Tuszynski, J.~A.
\newblock Entropic {Fragment}-{Based} {Approach} to {Aptamer} {Design}.
\newblock \emph{Chemical Biology \& Drug Design}, 78\penalty0 (1):\penalty0 1--13, 2011.
\newblock Publisher: Wiley Online Library.

\bibitem[Vaswani et~al.(2023)Vaswani, Shazeer, Parmar, Uszkoreit, Jones, Gomez, Kaiser, and Polosukhin]{vaswani_attention_2023}
Vaswani, A., Shazeer, N., Parmar, N., Uszkoreit, J., Jones, L., Gomez, A.~N., Kaiser, L., and Polosukhin, I.
\newblock Attention {Is} {All} {You} {Need}, August 2023.
\newblock URL \url{http://arxiv.org/abs/1706.03762}.
\newblock arXiv:1706.03762 [cs].

\bibitem[Wang et~al.(2022)Wang, Mistry, and Chou]{wang_discrete_2022}
Wang, Y., Mistry, B.~A., and Chou, T.
\newblock Discrete stochastic models of {SELEX}: {Aptamer} capture probabilities and protocol optimization.
\newblock \emph{The Journal of Chemical Physics}, 156\penalty0 (24), 2022.
\newblock Publisher: AIP Publishing.

\bibitem[Wang et~al.(2024)Wang, Liu, Zhang, Li, Feng, Lv, Diao, Luo, Yan, He, and {others}]{wang_aptadiff_2024}
Wang, Z., Liu, Z., Zhang, W., Li, Y., Feng, Y., Lv, S., Diao, H., Luo, Z., Yan, P., He, M., and {others}.
\newblock {AptaDiff}: de novo design and optimization of aptamers based on diffusion models.
\newblock \emph{Briefings in Bioinformatics}, 25\penalty0 (6):\penalty0 bbae517, 2024.
\newblock Publisher: Oxford University Press.

\bibitem[Zhang \& Sennrich(2019)Zhang and Sennrich]{zhang_root_2019}
Zhang, B. and Sennrich, R.
\newblock Root {Mean} {Square} {Layer} {Normalization}, October 2019.
\newblock URL \url{http://arxiv.org/abs/1910.07467}.
\newblock arXiv:1910.07467 [cs].

\bibitem[Zhang et~al.(2023)Zhang, Jiang, Kuster, Ye, Huang, Fürbacher, Zhang, Tang, Ibberson, Wild, and {others}]{zhang_single-step_2023}
Zhang, Y., Jiang, Y., Kuster, D., Ye, Q., Huang, W., Fürbacher, S., Zhang, J., Tang, Z., Ibberson, D., Wild, K., and {others}.
\newblock Single-step discovery of high-affinity {RNA} ligands by {UltraSelex}.
\newblock 2023.

\bibitem[Zhang et~al.(2024)Zhang, Chao, Jin, Zhang, Zhou, Yang, Yang, Huang, Yang, Xu, and {others}]{zhang_rnagenesis_2024}
Zhang, Z., Chao, L., Jin, R., Zhang, Y., Zhou, G., Yang, Y., Yang, Y., Huang, K., Yang, Q., Xu, Z., and {others}.
\newblock {RNAGenesis}: {Foundation} {Model} for {Enhanced} {RNA} {Sequence} {Generation} and {Structural} {Insights}.
\newblock \emph{bioRxiv}, pp.\  2024--12, 2024.
\newblock Publisher: Cold Spring Harbor Laboratory.

\bibitem[Zhao et~al.(2024)Zhao, Oono, Takizawa, and Kotera]{zhao_generrna_2024}
Zhao, Y., Oono, K., Takizawa, H., and Kotera, M.
\newblock {GenerRNA}: {A} generative pre-trained language model for de novo {RNA} design.
\newblock \emph{PLOS ONE}, 19\penalty0 (10):\penalty0 e0310814, October 2024.
\newblock ISSN 1932-6203.
\newblock \doi{10.1371/journal.pone.0310814}.
\newblock URL \url{https://dx.plos.org/10.1371/journal.pone.0310814}.

\end{thebibliography}
\bibliographystyle{icml2025}

\newpage
\onecolumn
\section*{Appendix}
\appendix

\counterwithin{figure}{section}
\counterwithin{table}{section}
\counterwithin{equation}{section}
\counterwithin{algorithm}{section}

\renewcommand{\thefigure}{\Alph{section}.\arabic{figure}}
\renewcommand{\thetable}{\Alph{section}.\arabic{table}}
\renewcommand{\theequation}{\Alph{section}.\arabic{equation}}
\renewcommand{\thealgorithm}{\Alph{section}.\arabic{algorithm}}

\section{RNA-BAnG Architecture}
\label{app:bang_model}

\begin{figure}[ht]
\vskip 0.2in
\begin{center}
\begin{minipage}[b]{0.40\columnwidth}
\centering
\includegraphics[width=\linewidth]{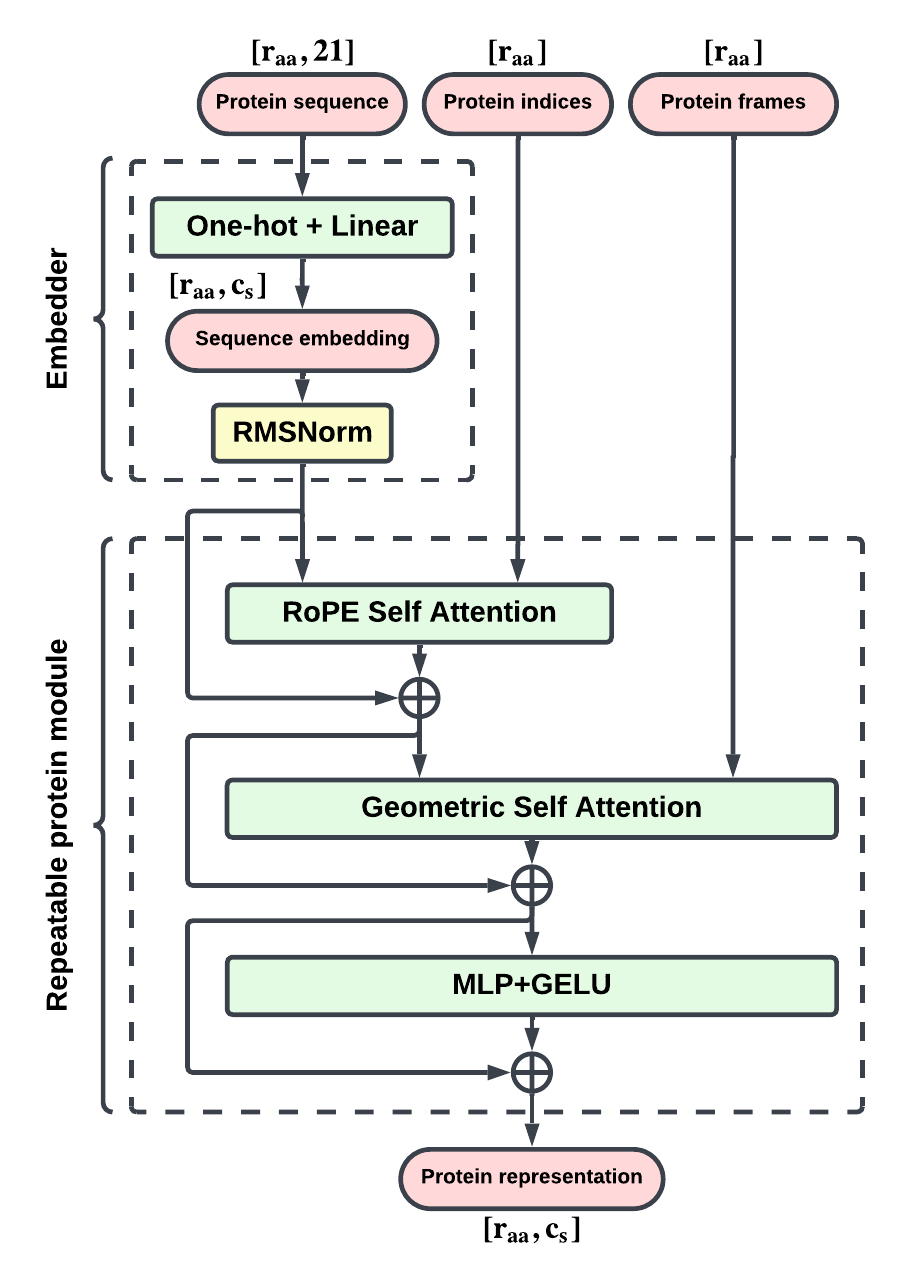}
\end{minipage}
\hfill
\begin{minipage}[b]{0.57\columnwidth}
\centering
\includegraphics[width=\linewidth]{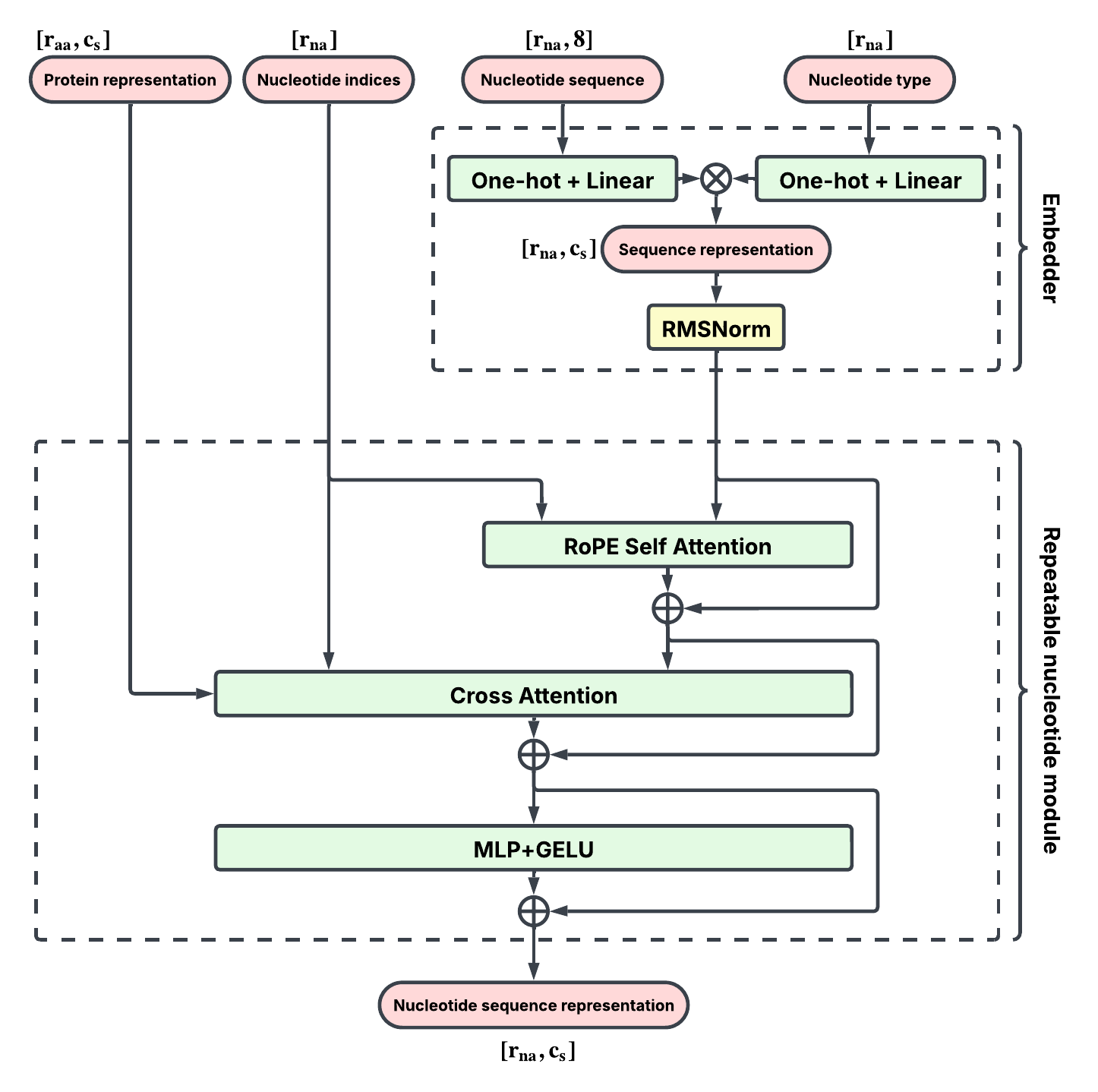}
\end{minipage}
\caption{Schematic illustration of the protein (on the left) and nucleotide (on the right) modules in the RNA-BAnG architecture. Here, ${\bf r_{na}}$ is the number of nucleotides, ${\bf r_{aa}}$ is the number of protein residues, $\bigotimes$ stands for concatenation.}
\label{fig:arc_modules}
\end{center}
\vskip -0.2in
\end{figure}

RNA-BAnG architecture is schematically presented in \cref{fig:model} of the main text and is composed of protein and nucleotide modules, detailed in \cref{fig:arc_modules}.
The latent dimension of the model is $c_s = 128$, all heads dimensions are set to $c_h = 64$. Feedforward blocks have a scaling factor $n = 2$. We set the number of protein and nucleotide modules to 10 each. While reducing this number led to poorer performance, increasing it did not yield any noticeable improvements.
We adjusted the rest of the hyperparameters by selecting the smallest model size combination that ensured stable convergence. Resulting model contains 14,5 million parameters.

\cref{alg:attention} describes the \textbf{Geometric Attention} block. 
It takes frames $T_i$ and single representation $s_i$ of every protein residue in the chain as inputs. 
The number of attention heads is $h = 12$, the number of query points is $N_\text{query points} = 4$,  the number of value points is $N_\text{value points} = 8$.  
The weight per head $\gamma^h \in \mathbb{R}$ is the softplus of a learnable scalar. 
$\omega$ is a weighting factor.
We adjusted the block's hyperparameters to align with the AlphaFold2 IPA choices.

\begin{algorithm}[h]
   \caption{Geometric Attention}
   \label{alg:attention}
\begin{algorithmic}
   \INPUT $s_i$, $T_i$ 
   \STATE $q_i^{hp}, k_i^{hp} \gets \text{LinearNoBias}(s_i) \hfill p \in [1,  N_\text{query points}]$  
   \STATE $v_i^{hp} \gets \text{LinearNoBias}(s_i) \hfill p \in [1,  N_\text{point values}]$
   \STATE $w \gets \sqrt{\frac{2}{9N_\text{query points}}}$
   \STATE $a_{ij}^h \gets \text{softmax} \left( - \frac{\gamma^h w}{2} \sum_p \|T_i \circ q_i^{hp} - T_j \circ k_j^{hp} \|^2 \right)$
   \STATE $o_i^{hp} \gets T_i^{-1} \circ \sum_j a_{ij}^h \left( T_j \circ v_j^{hp} \right)$
   \OUTPUT $\text{Linear} \left( \text{concat}_{h,p} (o_i^{hp}, \| o_i^{hp} \|) \right)$  
\end{algorithmic}
\end{algorithm}

\section{Synthetic Task}
\subsection{Toy Model}
\label{app:synth_model}
To avoid direct memorization on the synthetic task, we opted for a compact model, same for each tested method. Its final configuration is a two-block transformer (with a hidden dimension of 64) with the RoPE positional attention and two attention heads, resulting in 17k parameters. We trained the model for each method for 80k steps with a batch size of 8, using the Adam optimizer \cite{kingma_adam_2017} with the learning rate of 0.0001. During the inference, the length was fixed to 50 tokens for iterative methods.

\subsection{Synthetic Sampling}
\label{app:synth_sample}

During the inference, the sequence length was fixed to 50 tokens for iterative methods. For all the methods tested on the synthetic task, we sample tokens from their distributions using the top-k strategy with $k=4$.

\subsection{Generated Synthetic Sequences}
\label{app:synth_seqs}
\cref{table:synth_seqs} lists examples of sequences generated for the DoubleBind task with each tested approach.

\begin{table}[h]
\caption{Example of generated sequences for the synthetic task. Expected synthetic motifs are underlined.}
\label{table:synth_seqs}
\vskip 0.15in
\begin{center}
\begin{small}
\begin{sc}
\begin{tabular}{lcccr}
\toprule
Generative method & Sequences \\
\midrule
BAnG    &   \parbox{12cm}{ucccggcugguuccgaucggaac\underline{UGACUC}gcaccugguuccgacuuauau \\ uaucgguacgcuacaggcgucu\underline{CAAUUG}agagcggcuggcauguauugcu} \\
\midrule
autoregressive & \parbox{12cm}{ggac\underline{CAAUUG}uugau\underline{UGACUC}ggacuaauugcuccaccgcuaacgauugc} \\
\midrule
IAnG entropy  & \parbox{12cm}{ccu\underline{CAAUUG}gggagccggcugccgugcugcgagugacauuugaacgugau \\ cucgccgcccggggaccauugcacaaau\underline{UGACUC}uagcgcagaaugguac}\\
\midrule
iterative entropy  & \parbox{12cm}{auagaauuuacccaccugaugaugccccacuuagcggauaucugcuuucg \\ aaaauauucgugguuuuacuccaccacuccuaaacgcgaacugaaccuac} \\
\bottomrule
\end{tabular}
\end{sc}
\end{small}
\end{center}
\vskip -0.1in
\end{table}

\subsection{Additional Synthetic Results}
\label{app:synth_results}

To assess the robustness of BAnG to sequence length, we conducted additional synthetic experiments. Our method was tested on the SingleBind task under three scenarios: sequence lengths of 100 tokens, 200 tokens, and uniformly sampled between 40 and 50 tokens. The results, presented in \cref{table:syth_len}, show that BAnG performs consistently across these variations. The average length of the generated sequences was 46 tokens when trained on sequences sampled from the 40-50 token range.

\begin{table}[ht]
\caption{BAnG performance on SingleBind task with various sequence length. Scores represent the proportion of sequences that contain the correct synthetic motif.}
\label{table:syth_len}
\vskip 0.15in
\begin{center}
\begin{small}
\begin{sc}
\begin{tabular}{lcccr}
\toprule
Length & Score $\uparrow$\\
\midrule
50    & {\bf 0.98}  \\
100 & {\bf 0.98}  \\
200  & 0.95  \\
$\mathcal{U}(40,50)$  & {\bf 0.98} \\
\bottomrule
\end{tabular}
\end{sc}
\end{small}
\end{center}
\vskip -0.1in
\end{table}

We also investigated BAnG's behavior when multiple motifs are present simultaneously in a sequence. DualBindMix is a variation of the DualBind task in which two motifs, separated by three tokens, appear in every training sequence, with anchor points placed randomly in the middle of either motif. The task, where motif separation is random, is referred to as DualBindMix Random. The results of these experiments are listed in \cref{table:syth_mixing}. In the DualBindMix BAnG generated the sequences containing two motifs simultaneously, while in the DualBindMix Random most of the time it generated only one motif. As expected, only the increased uncertainty in motif positioning negatively impacts the performance of our method.

\begin{table}[ht]
\caption{BAnG performance on tasks with several synthetic motifs. Values represent the proportion of sequences containing either one of the synthetic motifs or both. }
\label{table:syth_mixing}
\vskip 0.15in
\begin{center}
\begin{small}
\begin{sc}
\begin{tabular}{lcccr}
\toprule
Task & First & Second & Both \\
\midrule
DualBind    & 0.43 & 0.53 & 0.01 \\
DualBindMix & 0.01 & 0.01 & 0.97 \\ 
DualBindMix Random & 0.47 & 0.46 & 0.05 \\
\bottomrule
\end{tabular}
\end{sc}
\end{small}
\end{center}
\vskip -0.1in
\end{table}

\section{Training Data}
\subsection{Protein Coupled Nucleotide Sequences}
\label{app:data_double}
We defined two protein-RNA or protein-DNA chains as interacting if at least one interaction occurred between their residues. We defined an interaction between two residues if they share at least one atom-atom contact, as calculated using the VoroContacts software \cite{olechnovic_vorocontacts_2021}. The VoroContacts method identifies contacts based on the Voronoi tessellation of atomic balls, constrained within the solvent-accessible surface \cite{olechnovic_voronota_2014}.

We excluded all samples where protein atom coordinates are ambiguous (had alternative locations), where the protein chain contains non-canonical residues (though we substitute 'SCE' with 'CYS' and 'MSE' with 'MET' beforehand), or where the nucleotide chain contains non-standard residues or a mixture of standard RNA and DNA residues. We also excluded samples with the nucleotide sequence length of less than 10 residues. To avoid potential computational resource problems, we included in the training only samples with a protein length of fewer than 500 residues.

During training, we split the data by protein sequence homology. We measured it using the clustering provided by PPI3D \cite{dapkunas_ppi3d_2024}, where proteins within the same cluster share less than 40\% sequence similarity with those in other clusters.  We allocated samples from 95\% of randomly selected clusters to the train set and the rest we used for validation. This approach allows tracking the model generalization across the protein space. To enhance our potential test set, we removed from the training data clusters containing proteins from the PDB samples 5ITH (chain A), 7CRE (chain A), 6QW6 (chain R), 8OPS (chain B), and 1CVJ (chain A).

Some proteins and nucleotide sequences are overrepresented in the data, which may lead to training imbalance. To address this, we introduced an additional level of sample clustering. First, we clustered nucleotide sequences (DNAs and RNAs separately) based on sequence similarity, grouping those with 90\% identity using CD-HIT-EST \cite{li_cd-hit_2006}. Then, we clustered the samples based on the combination of protein and nucleotide sequences clusters. At each epoch during training, we selected eight random samples (the mean cluster population) to represent each of the latter clusters, repeating the samples when necessary.

The resulting dataset consisted of 123,043 samples, distributed across 3,580 protein sequence clusters, 2,807 nucleotide sequence clusters (915 RNA and 1,892 DNA), and 12,667 combined clusters. The protein sequences in the dataset have a mean length of 155 residues with a standard deviation of 90. For nucleotide sequences, RNA lengths average 1,834 nucleotides ($\pm$ 1,564), while DNA lengths average 76 nucleotides ($\pm$ 78).

To prevent potential computational resource issues and to focus the model on the binding motifs, we truncated nucleotide sequences exceeding 300 residues during training and validation. This truncation limited sequences to 300 residues centered around the anchor point, which was selected as described prior to the truncation.

\subsection{Standalone RNA Sequences}
\label{app:data_solo}
From RNAcentral database we selected sequences containing only standard residues and with lengths between 10 and 500 nucleotides. The selection was then deduplicated using CD-HIT-EST with a 90\% similarity threshold, resulting in a final set of 3 million sequences.

\section{Training Details}
\label{app:train}
As explained in the main text, we conducted the RNA-BAnG training in two steps.
In the first step, we used a batch size of 64 and trained for 255k steps. In the second step, we used a batch size of 8 and trained for 216k steps. We stopped training when the validation loss decline became visually negligible. In both training steps, we used the ASMGrad \cite{reddi_convergence_2019} variation of the Adam optimizer \cite{kingma_adam_2017} with default parameters and the learning rate of 0.0001. Learning rate was warmed up linearly for the first 1k steps and then decayed exponentially with $\gamma = 0.99$ and a period of 1k steps. 
Complete training took 4 days on a single MI120 AMD GPU.

\section{Evaluation Protocol}
\subsection{Test Data}
\label{app:eval_test}

The RNA Compendium study provides 244 samples, each comprising a protein sequence paired with approximately 200,000 RNA sequences and their corresponding experimental binding scores. 
These samples are designated by the authors as RNCMPT00XXX, where XXX represents a numerical identifier. 
For simplicity, we occasionally refer to these samples solely by their numerical identifiers XXX.

We processed the data as follows. For each sample, RNA sequences were ranked by their binding scores. The top 2,000 sequences were labeled as the positive (interacting) class, while the bottom 2,000 were labeled as the negative (non-interacting) class. To eliminate sequence redundancy and prevent data leakage, duplicate RNA sequences were removed using CD-HIT-EST with a sequence identity threshold of 90\%. The remaining sequences were then randomly split into training and testing sets, so that each test set would have a 1,000 sequences. Those test sets are referenced as positive and negative sets in the main manuscript. 
We then trained individual DeepCLIP models for each sample, following the protocol outlined by the authors. The DeepCLIP model contains approximately 3,000 trainable parameters, and its training protocol -- designed by the authors -- automatically selects the best-performing weights when provided with sets of positive and negative sequences.
Only models achieving an area under the receiver operating characteristic curve (AUROC) of 0.95 or higher on the corresponding test sets were selected for further analysis.

\subsection{RNA-BAnG Sampling}
\label{app:eval_sample}

For RNA-BAnG, we sample tokens from their distributions using the top-k strategy with $k=4$. Maximum sequence length is set to 50 nucleotides. Average sampling time is 20 minutes for 1,000 RNA sequences. To generate a random nucleotide sequence, we first choose a sequence length at random between 40 and 50 (inclusive). Then, we create a sequence of that length by selecting each nucleotide independently and uniformly at random. The length choice ensures that the mean sequence length of random samples matches that of RNA-BAnG-generated sequences (45.6 nucleotides).

\section{Evaluation Results and Comparative Approaches}
\label{app:results}

\subsection{Additional Results}

\cref{table:bang_seqs} lists randomly selected examples of generated RNA-BAnG sequences for the two best and two worst performance test samples.

\begin{table}[t]
\caption{Example of sequences generated by RNA-BAnG. Samples are ordered by decreasing proportion of high-affinity sequences. First two generated nucleotides are in bold and marked by red color.}
\label{table:bang_seqs}
\vskip 0.15in
\begin{center}
\begin{small}
\begin{sc}
\begin{tabular}{lcccr}
\toprule
RNAcompete ID & Sequences \\
\midrule
RNCMPT00077    &   \parbox{12cm}{AUUUUUAAAUAUU\textcolor{red}{\textbf{UU}}UAAAAAAAA \\ UAUU\textcolor{red}{\textbf{AU}}UUAUUUUAAAA \\ GUAUGUAUUUAU\textcolor{red}{\textbf{UU}}U} \\
\midrule
RNCMPT00173 & \parbox{12cm}{GUGA\textcolor{red}{\textbf{AC}}GUAAAACUUUUAACUUAAAUUCCUCA \\ AUUGAAA\textcolor{red}{\textbf{GC}}UUUUAUGCCUUUUACAAUAAA \\ CGACUCAAAAGACAAUCUAAUACU\textcolor{red}{\textbf{CA}}AAAACCGGAUUAAACUUAAAAAUA} \\
\midrule
RNCMPT00167  & \parbox{12cm}{CUUGUCU\textcolor{red}{\textbf{GA}}CACG \\ GCCCCUUGACCUUGAGUCCCAUGU\textcolor{red}{\textbf{GG}}CAGAGCAGUACAGGCUGAGUCGCU \\ UG\textcolor{red}{\textbf{AG}}GUACACCA} \\
\midrule
RNCMPT00133  & \parbox{12cm}{GCGCAGUGCCCAUAGACUCUGCAU\textcolor{red}{\textbf{AA}}UGGGACUCCAAGGAGCCGUCGGUU \\ CCUGCGAACUUAUCAUUUCUAUAG\textcolor{red}{\textbf{UG}}AUGCAAAUAUGUACUUAAUUUUUA \\ CGGAACGGAUUAUUUGUUUUAUAA\textcolor{red}{\textbf{AU}}AAUUAUGAAAAGUAUUUUAUUAUA}\\
\bottomrule
\end{tabular}
\end{sc}
\end{small}
\end{center}
\vskip -0.1in
\end{table}

\begin{figure}[ht]
\vskip 0.2in
\begin{center}
\begin{minipage}[b]{0.48\columnwidth}
\centering
\includegraphics[width=\linewidth]{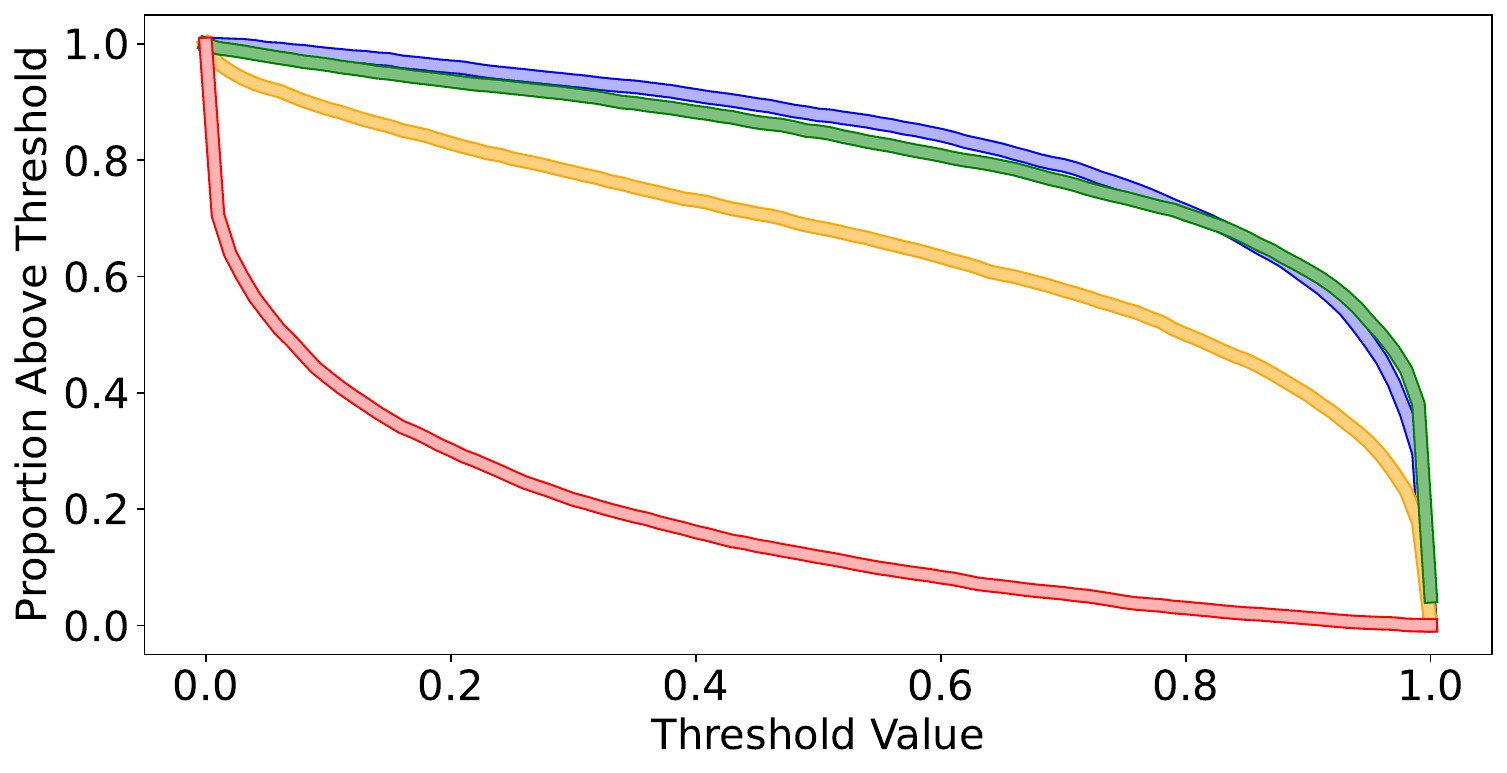}
\end{minipage}
\hfill
\begin{minipage}[b]{0.48\columnwidth}
\centering
\includegraphics[width=\linewidth]{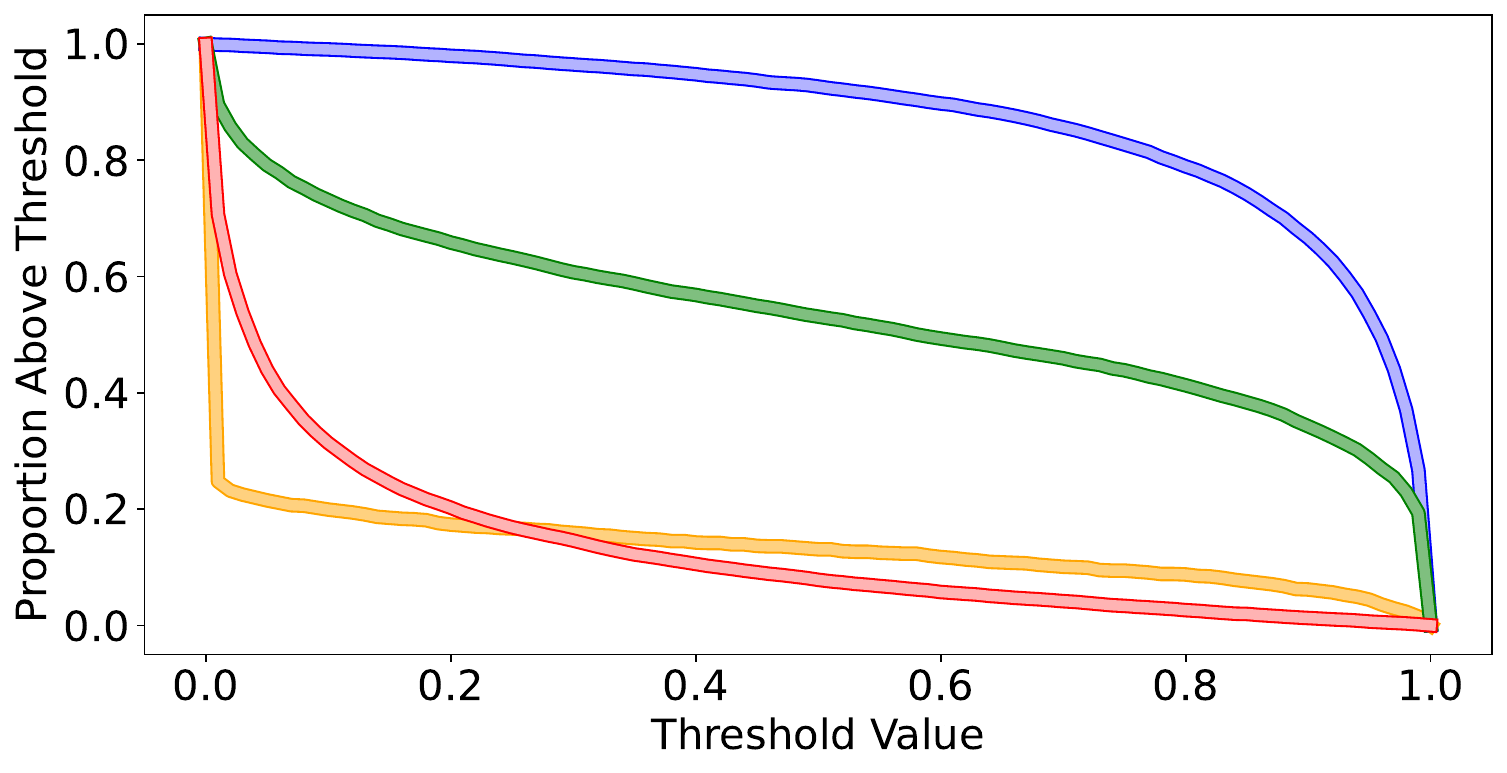}
\end{minipage}
\caption{Proportion of sequences above the threshold: The left panel shows results for GenerRNA (yellow), RNA-BAnG (green), and the corresponding positive (blue) and negative (red) experimental sets used in the GenerRNA vs. RNA-BAnG comparison. The right panel presents RNAFLOW (yellow), RNA-BAnG (green), and the positive (blue) and negative (red) experimental sets used in the RNAFLOW vs. RNA-BAnG comparison. The values represent averages over the entire comparison set in each case.}
\label{fig:cdf_models}
\end{center}
\vskip -0.2in
\end{figure}

We have also compared areas under the threshold-dependent performance curves for GenerRNA and RNAFLOW on corresponding comparison sets (curves shown in the \cref{fig:cdf_models}). RNA-BAnG demonstrated area values of 0.81 and 0.53, surpassing GenerRNA with 0.66 and RNAFLOW with 0.13.

\begin{figure}[ht]
\vskip 0.2in
\begin{center}
\centerline{\includegraphics[width=0.4\columnwidth]{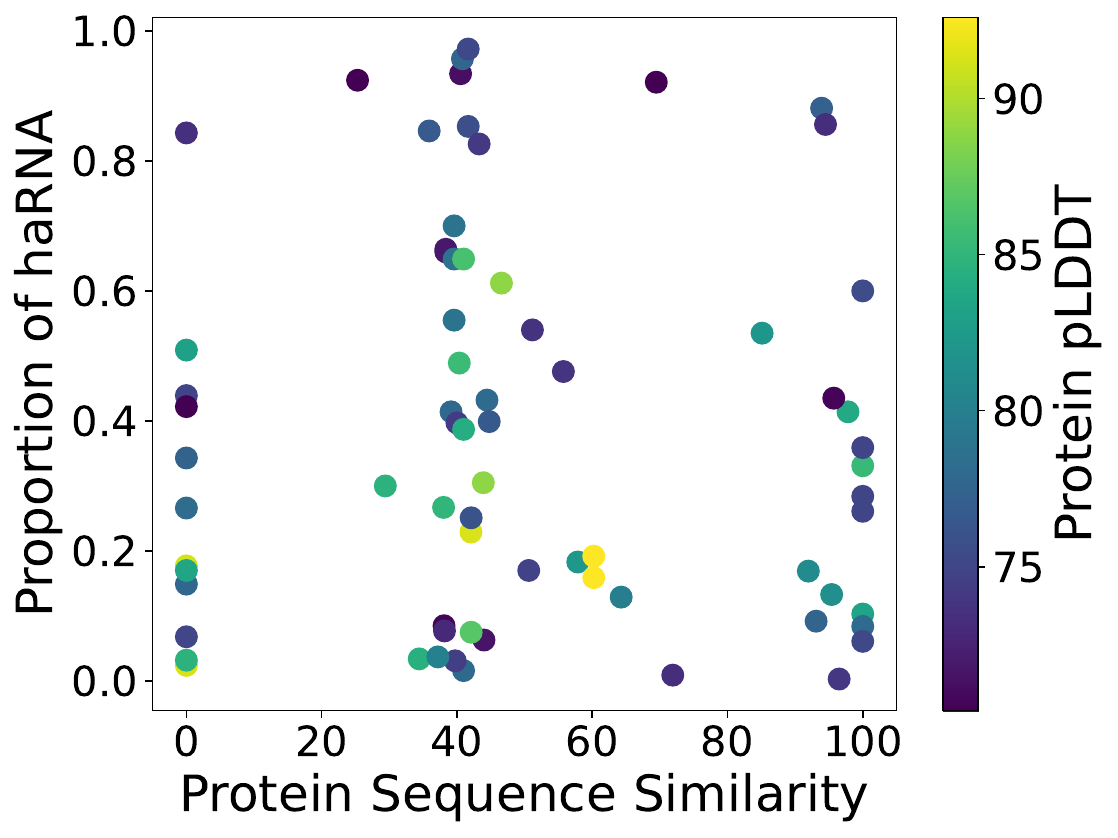}}
\caption{Proportion of generated high-affinity sequences as a function of protein sequence similarity to the training data and the AlphaFold pLDDT scores of the predicted protein structures.}
\label{fig:corr}
\end{center}
\vskip -0.2in
\end{figure}

\cref{fig:corr} shows the proportion of high-affinity generated sequences as a function of AlphaFold pLDDT scores and protein sequence similarity with the train set.

\begin{figure}[ht]
\vskip 0.2in
\begin{center}
\centerline{\includegraphics[width=\columnwidth]{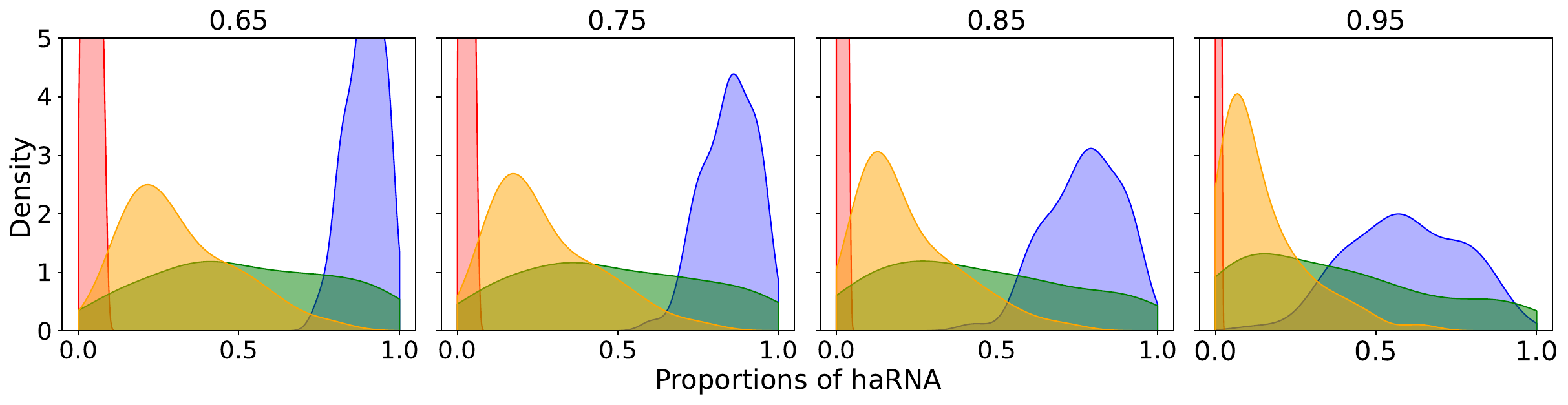}}
\caption{Density plots of the proportion of high-affinity sequences for several threshold values (0.65, 0.75, 0.85, and 0.95, listed on top) across the whole test set. RNA-BAnG is shown in green,  random sequences in yellow, positive experimental sequences in blue and negative experimental sequences in red.
}
\label{fig:res_dense}
\end{center}
\vskip -0.2in
\end{figure}

\cref{app:tools} details protein sequence similarity calculations. \cref{fig:res_dense} shows kernel density plots calculated over the whole test set of 71 samples. We assessed four different DeepCLIP score threshold values (0.65, 0.75, 0.85, 0.95, from left to right in the Figure). We can conclude that the value of 0.95 is too stringent, as the mean of the positive experimental sets approaches the value of 0.5. Subjectively, the most visually appealing threshold value is 0.75. Nonetheless, in the main text (\cref{fig:cdf}) we continuously assess all the values by plotting the mean proportions above a certain threshold value.

\begin{figure}[ht]
\vskip 0.2in
\begin{center}
\centerline{\includegraphics[width=\linewidth]{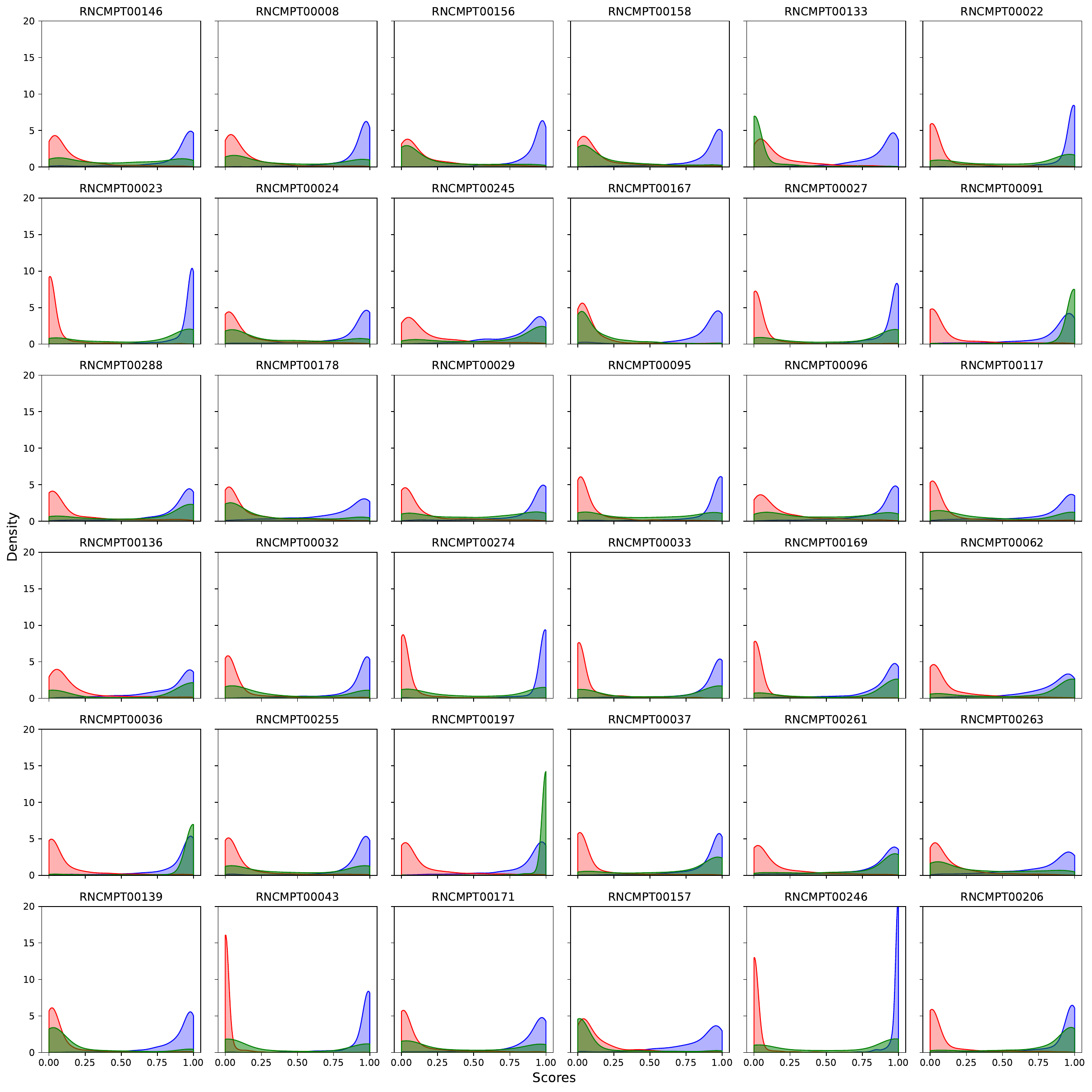}}
\caption{Distribution of DeepCLIP scores for the first part of the test samples (RNAcompete sample IDs on top). Scores of RNA-BAnG are in green. Scores of positive and negative experimental sequences are in blue and red, respectively.}
\label{fig:dens1}
\end{center}
\vskip -0.2in
\end{figure}

\begin{figure}[ht]
\vskip 0.2in
\begin{center}
\centerline{\includegraphics[width=\linewidth]{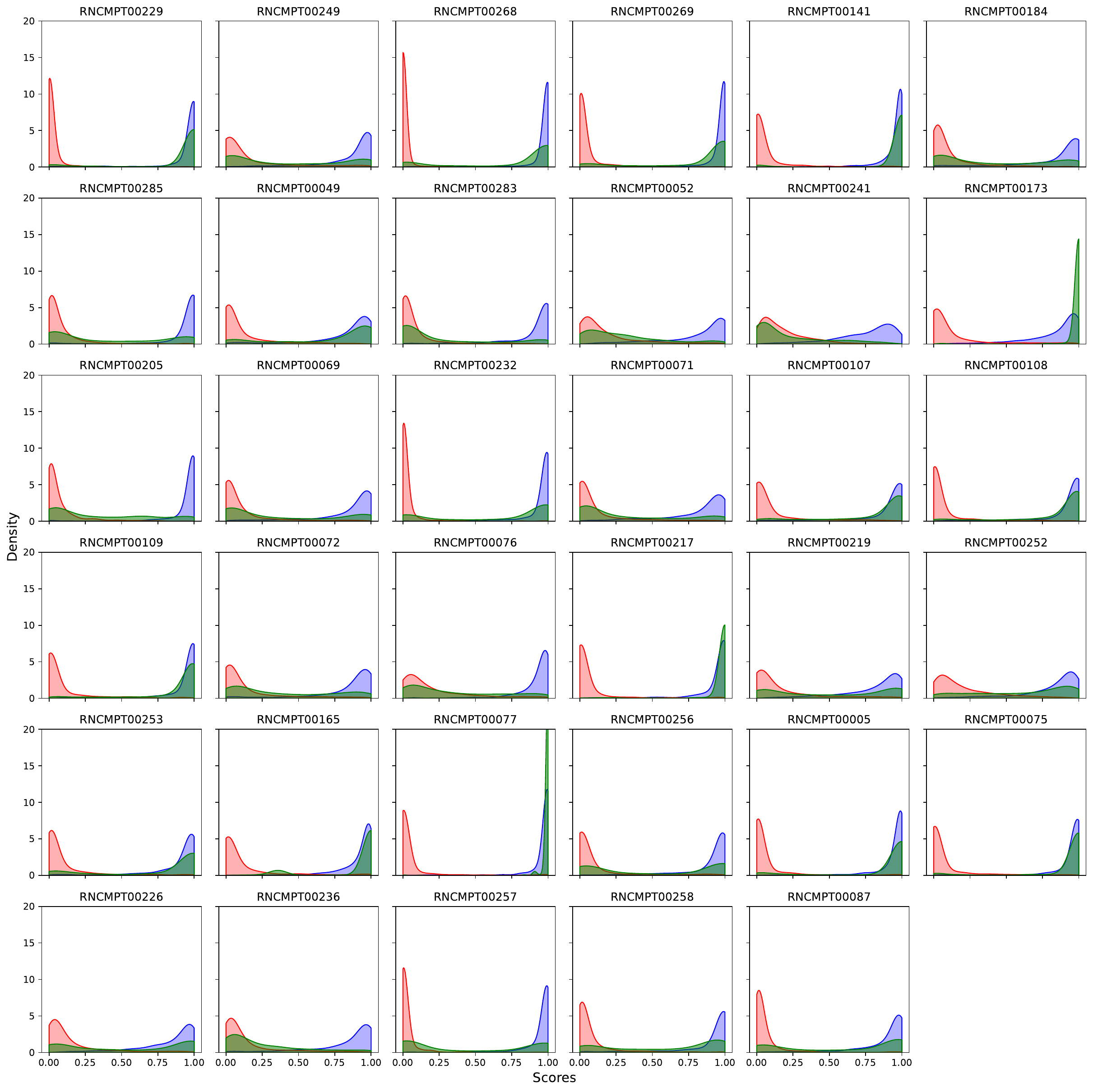}}
\caption{Distribution of DeepCLIP scores for the second part of test samples (RNAcompete sample IDs on top). Scores of RNA-BAnG are in green. Scores of positive and negative experimental sequences are in blue and red, respectively.}
\label{fig:dens2}
\end{center}
\vskip -0.2in
\end{figure}

Distributions of DeepCLIP scores for each of 71 test samples are depicted in \cref{fig:dens1} and \cref{fig:dens2}.

\subsection{Ablation Studies}

To better understand the contributions of individual components in our model, we conducted ablation studies focusing on {\bf Geometric Attention} and the use of relative position encodings in {\bf Cross Attention}. Removing the {\bf Geometric Attention} block resulted in a relatively small change in the number of model parameters (approximately 10\%), while removing relative position encodings led to a change of less than 1\%.

\begin{figure}[ht]
\vskip 0.2in
\begin{center}
\begin{minipage}[b]{0.48\columnwidth}
\centering
\includegraphics[width=\linewidth]{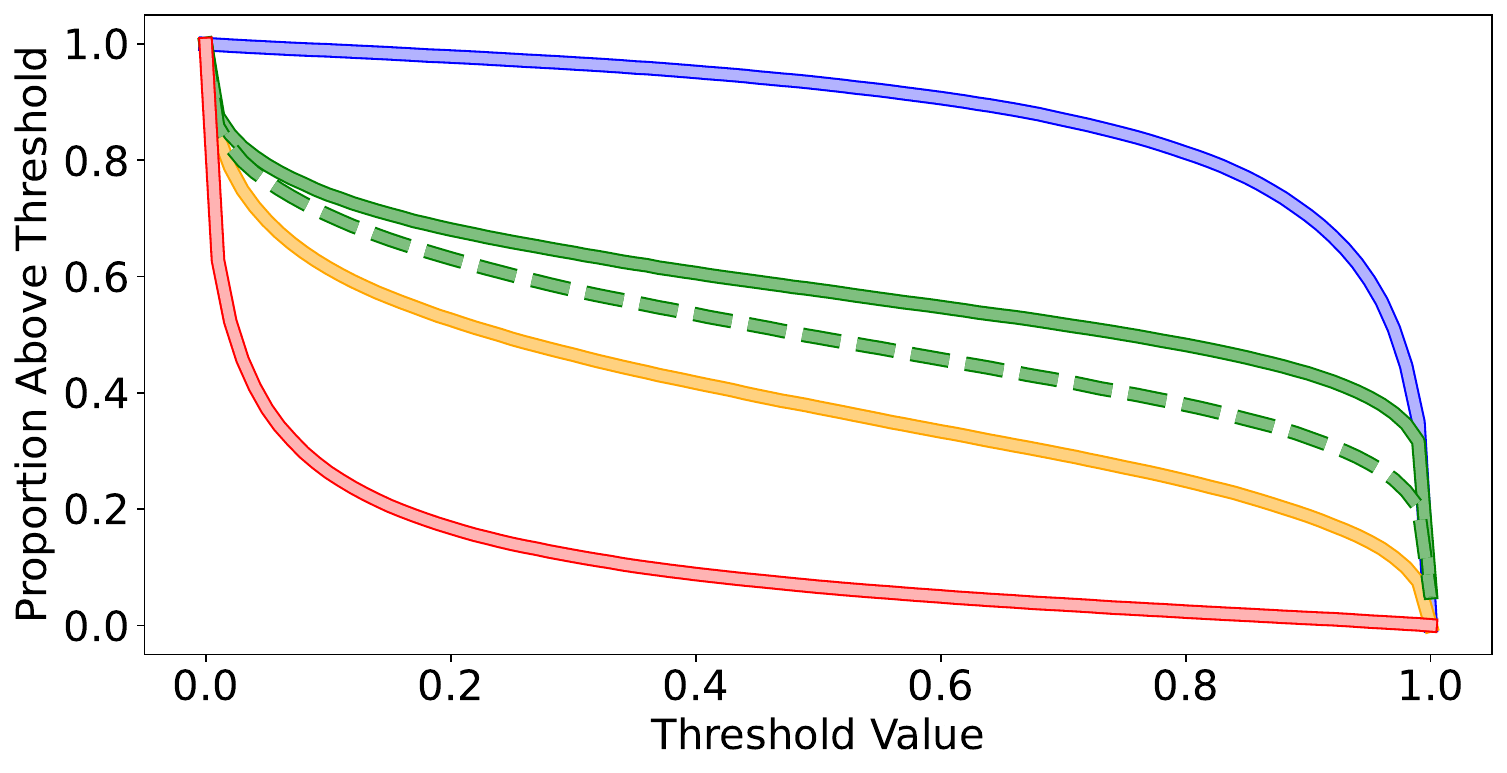}
\end{minipage}
\hfill
\begin{minipage}[b]{0.48\columnwidth}
\centering
\includegraphics[width=\linewidth]{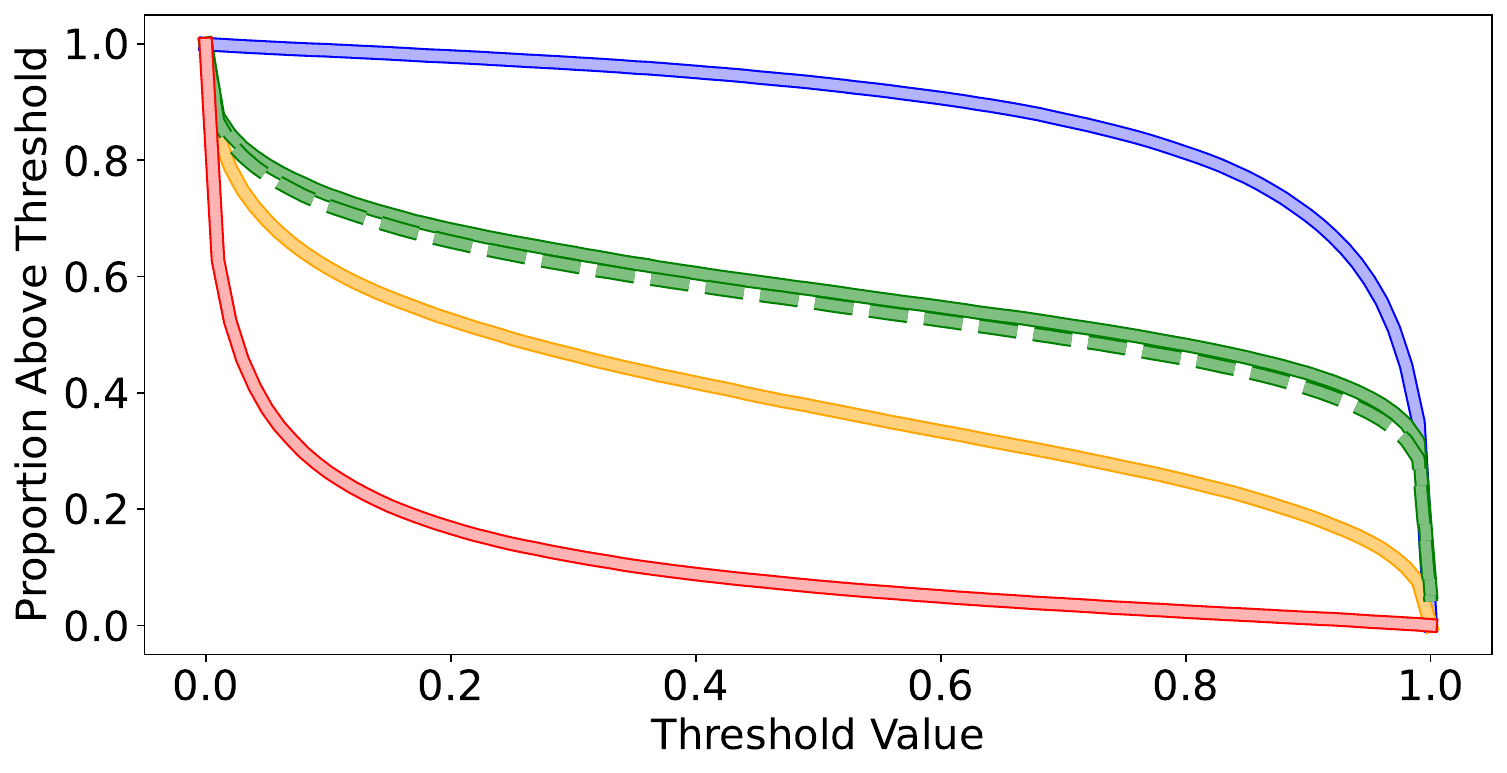}
\end{minipage}
\caption{Proportion of sequences above the threshold: generated by RNA-BAnG (green) and randomly (yellow), the positive (blue) and negative (red) experimental sets. Dashed green line corresponds to results of RNA-BAnG generation without {\bf Geometric Attention} (on the left) or without relative position encodings in {\bf Cross Attention} (on the right). The values here represent the averages for the entire comparison set.}
\label{fig:cdf_ablation}
\end{center}
\vskip -0.2in
\end{figure}

As shown in \cref{fig:cdf_ablation}, {\bf Geometric Attention} significantly enhances the performance of RNA-BAnG. While the contribution of relative position encodings in {\bf Cross Attention} is more modest, it still provides a measurable improvement.

\subsection{Protein Preprocessing}

\begin{figure}[ht]
\vskip 0.2in
\begin{center}
\centerline{\includegraphics[width=0.7\linewidth]{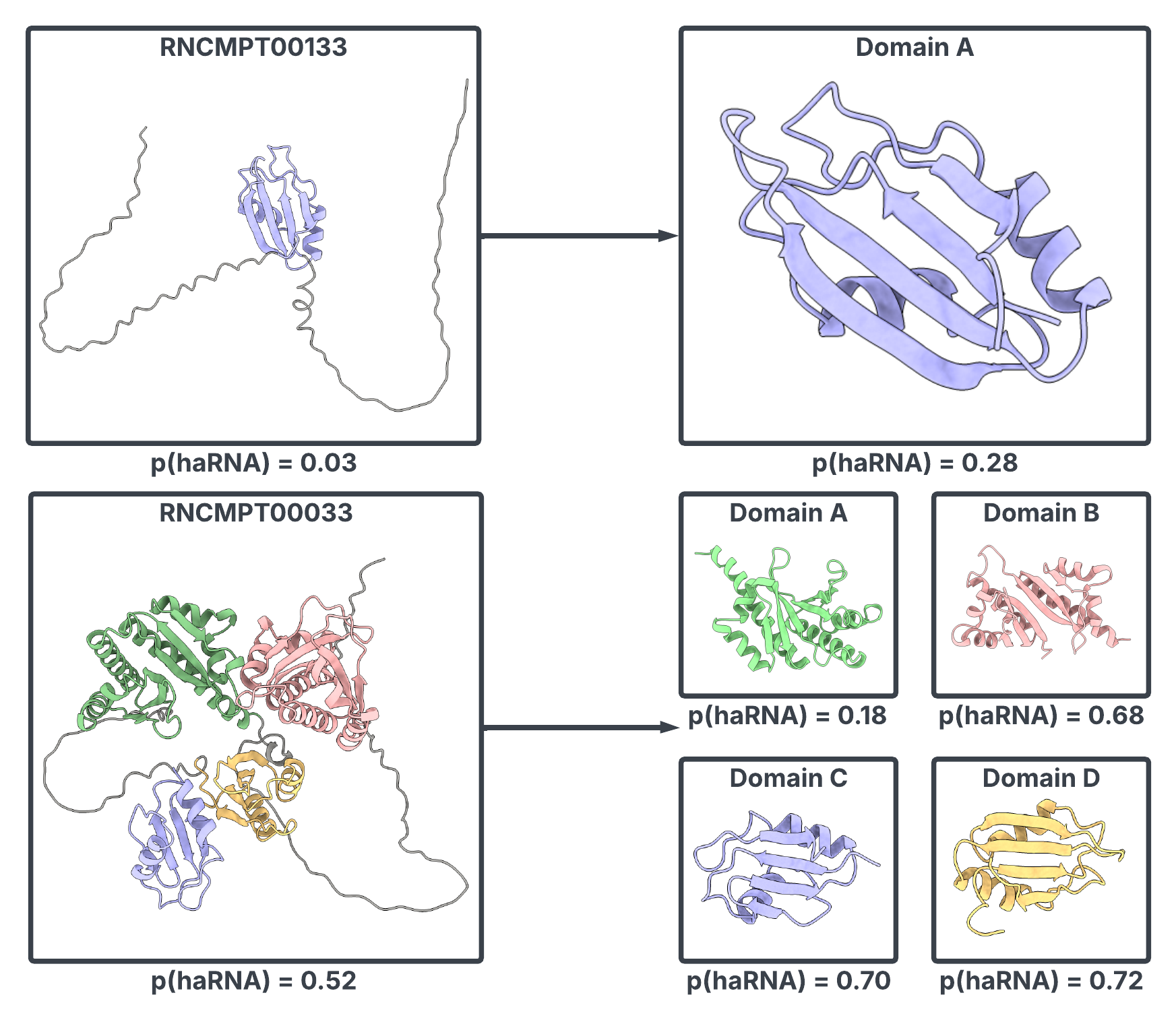}}
\caption{Proteins before (on the left) and after (on the right) the preprocessing. The proportion of haRNA generated by RNA-BAnG is reported below the proteins or their domains . The domains were split using visual inspection guided by the pLDDT confidence scores.}
\label{fig:preproc}
\end{center}
\vskip -0.2in
\end{figure}

We also investigated the impact of protein data preprocessing on RNA-BAnG performance. Our findings suggest that segmenting the protein structure into individual domains or removing intrinsically disordered regions (IDRs) can, in some cases, enhance the quality of generated sequences (\cref{fig:preproc}). By domain here, we mean a compact, spatially distinct region of the protein with an ordered structure — recognized visually, by simply inspecting the structure and seeing how it can be split into distinct rigid blocks. 

For instance, in the case of the protein RNCMPT00133 (RNAcompete code), where RNA-BAnG performs the worst, removing IDRs (according to AlphaFold2 pLDDT scores) from the structure led to a tenfold increase in the number of generated high-affinity RNAs. Another example is protein RNCMPT00033, whose AlphaFold2-predicted structure contains multiple domains -- running the model on individual domains revealed that some domains yield significantly better results than using the full-length protein. However, such preprocessing is generally only practical when meta-information about protein–RNA interactions is available, as RNAs may interact with multiple domains or IDRs simultaneously.

\subsection{Alternative Sequence Similarity Scoring Approach}

\begin{figure}[H]
\vskip 0.2in
\begin{center}
\centerline{\includegraphics[width=0.8\linewidth]{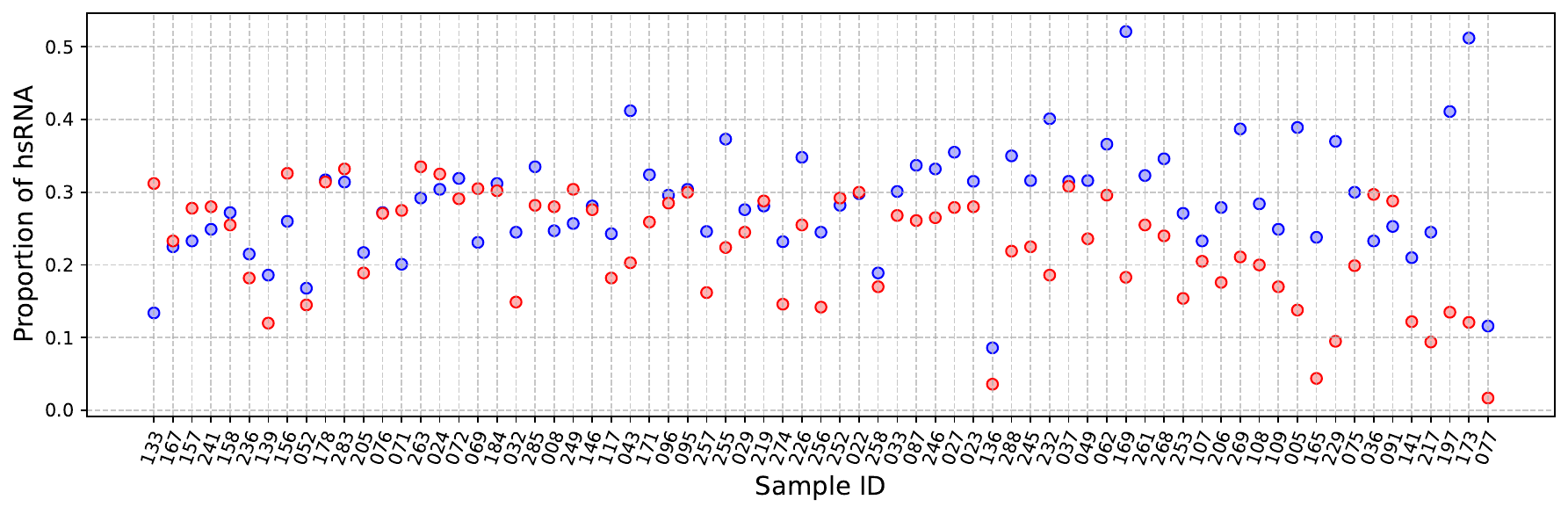}}
\caption{Proportions of high-similarity RNA sequences, generated by RNA-BAnG, to positive (blue) and negative (red) experimental sets for each test sample. Test samples are ordered by proportion of haRNA sequences, as in \cref{fig:res}. Sample IDs mapping to RNAcompete IDs mentioned in \cref{app:eval_test}}
\label{fig:exp_sim}
\end{center}
\vskip -0.2in
\end{figure}

We calculated the proportions of highly similar sequences (hsRNA) between the generated and experimental sets. A sequence is considered highly similar if identified as such by blastn with \texttt{-task blastn} parameter. The resulting values for each test sample are shown in \cref{fig:exp_sim}. Across different samples, the proportions of generated sequences similar to the positive set are higher than those of similar to the negative set, averaging 29\% and 23\%, respectively. However, it is important to note the limited applicability of blastn in cases where sequences are intentionally designed without evolutionary signals and are experimentally selected based on the presence of relatively short binding motifs \cite{ray_compendium_2013}.

\subsection{Alternative Motif Search Scoring Approach}

Motif searching tools in bioinformatics are specialized software applications designed to identify recurring patterns, or motifs, within biological sequences such as DNA, RNA, or proteins. Motifs are usually represented as Position Weight Matrices (PWMs), which capture the variability and conservation of each nucleotide (or amino acid) at every position within the motif \cite{stormo_use_1982}. PWMs are obtained by aligning multiple sequences that share the motif and calculating the frequency of each symbol at each position, often followed by converting these frequencies into probabilities or log-odds scores \cite{schneider_sequence_1990}. Once constructed, PWMs are then used to scan larger sequences or entire genomes to predict new sites with similar patterns. 

As mentioned previously, the sequences in our experimental set are intentionally designed without evolutionary signals, making them inherently random in nature. This randomness makes aligning the sequences problematic. Therefore, we followed the approach used by the RNA Compendium authors, considering only the top-10 scoring 7-mers (scoring data provided by the authors). We then aligned these 7-mers using MAFFT \cite{katoh_mafft_2013}, removing columns with gaps in more than half of the rows. This removal occurred only at the beginning or end of the alignment, ensuring no changes to the core motifs. The resulting aligned 7-mers were used to construct PWMs for each sample where the data was available in our experimental set.

\begin{figure}[H]
\vskip 0.2in
\begin{center}
\centerline{\includegraphics[width=0.5\linewidth]{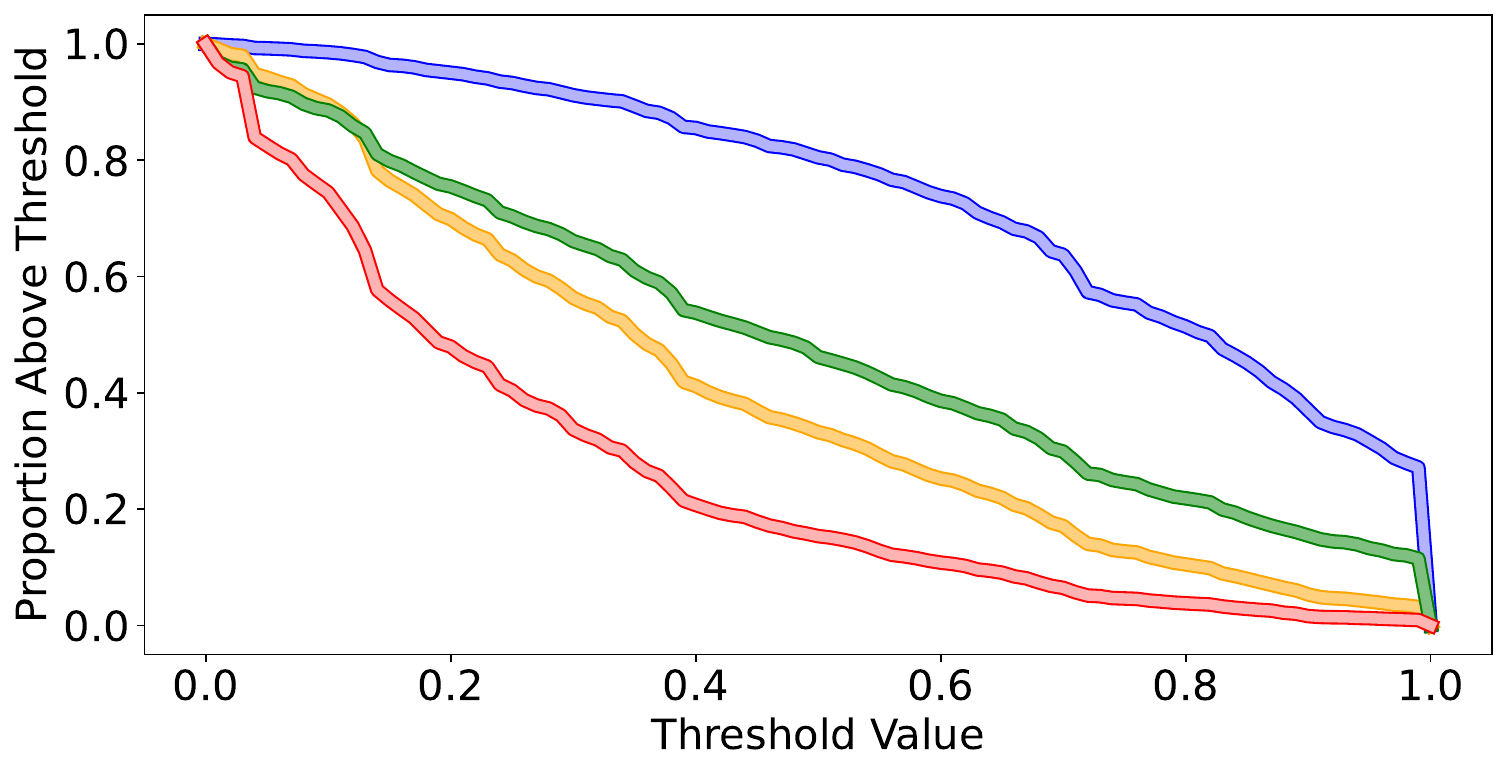}}
\caption{Proportion of sequences above the threshold: generated by RNA-BAnG (green) and randomly (yellow), the positive (blue) and negative (red) experimental sets. The values here represent the averages for the entire test set.}
\label{fig:cdf_thresh_moods}
\end{center}
\vskip -0.2in
\end{figure}

\begin{figure}[H]
\vskip 0.2in
\begin{center}
\centerline{\includegraphics[width=0.8\linewidth]{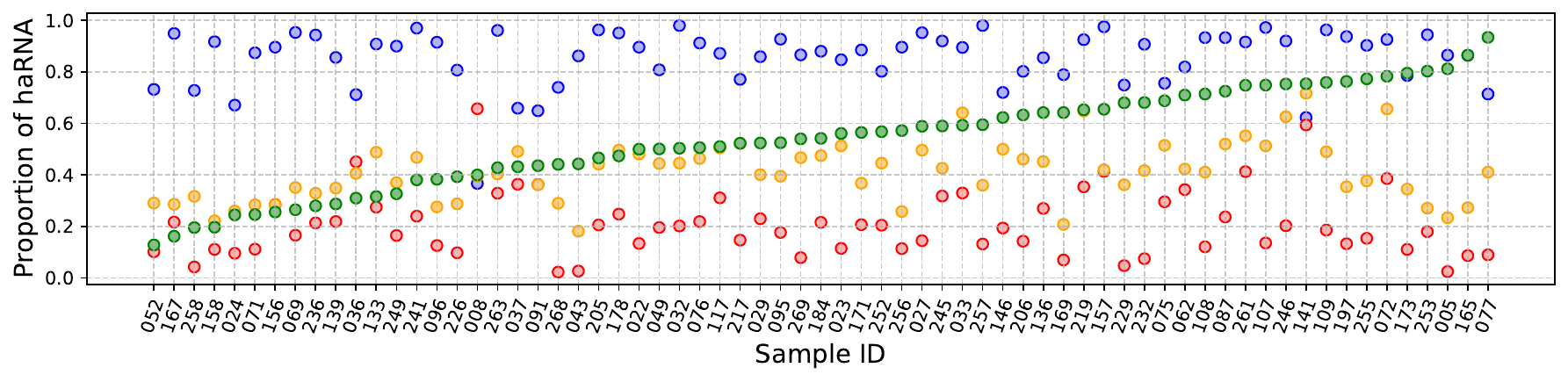}}
\caption{Proportions of high-affinity RNA sequences for each test sample, generated by RNA-BAnG (green) and randomly (yellow). These are compared with the proportions in the positive (blue) and negative (red) experimental sets. Test samples are ordered by RNA-BAnG performance. Sample IDs mapping to RNAcompete IDs mentioned in \cref{app:eval_test}.}
\label{fig:results_moods}
\end{center}
\vskip -0.2in
\end{figure}

Next, we used MOODS \cite{korhonen_fast_2017} as the motif search tool. We selected it for its ease of use and its ability to incorporate dinucleotide frequencies, which improves search accuracy. We mapped its scores to a 0-1 range using a sigmoid function. Similar to the approach with the DeepCLIP scores, Figure \cref{fig:cdf_thresh_moods} presents a threshold-dependent performance curve, showing the relationship between sequences with MOODS (sigmoid) scores above a certain threshold and the threshold value itself. A high area value of 0.74 for the positive experimental set, coupled with a very low area value of 0.26 for the negative set, indicates that motif-based scoring is effective at distinguishing high-affinity sequences from others. RNA-BAnG achieved a value of 0.49, indicating moderate success in generating sequences that partially align with the positive set and demonstrating some ability to avoid producing low-affinity outputs. The random set, with an area value of 0.40, suggests that our model outperforms random generation. To provide a more detailed analysis, we defined haRNAs as those in which the highest motif score exceeds 0.4, a threshold selected using the elbow method from \cref{fig:cdf_thresh_moods}. The resulting sample-by-sample statistics are shown in \cref{fig:results_moods}.

This scoring approach correlates with the one based on DeepCLIP scores, showing a Pearson correlation coefficient of 0.64 between the proportion of haRNA sequences defined by DeepCLIP and those defined by motif-based scoring for each sample. Additional details on the usage of MAFFT and MOODS can be found in \cref{app:tools}. 

\section{Alignment Tools Parameters}
\label{app:tools}
Clustering of nucleotide sequences was always performed using CD-HIT-EST \cite{li_cd-hit_2006} with a sequence identity threshold of 90\%. We used CD-HIT-EST with the following parameters: \texttt{-c 0.9 -n 9 -d 0 -T 10 -U 10 -l 9}. For protein sequence identity to the train set calculations we used blastp \cite{altschul_basic_1990} with default parameters. For the nucleotide sequences similarity to the train set calculations we used blastp \cite{altschul_basic_1990} with default parameters.

For the 7-mers alignment we used MAFFT \cite{katoh_mafft_2013} with the \texttt{--nuc} parameter. Motif search was performed with MOODS \cite{korhonen_fast_2017} using these parameters: \texttt{-R -p 1.0 --batch}. Motifs were represented in \texttt{adm} format. We have also set search window size equal to the motif size. 

\end{document}